# Linking heterogeneous microstructure informatics with expert characterization knowledge through customized and hybrid vision-language representations for industrial qualification

Mutahar Safdar[1,2], Gentry Wood[3], Max Zimmermann[4], Guy Lamouche[2], Priti Wanjara[2], Yaoyao Fiona Zhao[1,*]

[1]Department of Mechanical Engineering, McGill University, Montreal, QC, H3A 0C3, Canada
[2]National Research Council Canada, Montreal, QC, H3T 1J4, Canada
[3]Apollo-Clad Laser Cladding, a division of Apollo Machine and Welding Ltd., Edmonton, AB, T6E 5V2, Canada
[4]Fraunhofer Institute for Laser Technology ILT, Aachen, 52074, Germany
*Corresponding author: **yaoyao.zhao@mcgill.ca** | +1 (514) 398-2523

**Abstract**

Rapid and reliable qualification of advanced materials remains a bottleneck in industrial manufacturing, particularly for heterogeneous structures produced via non-conventional additive manufacturing processes. This study introduces a novel framework that links microstructure informatics with a range of expert characterization knowledge using customized and hybrid vision-language representations (VLRs). By integrating deep semantic segmentation with pre-trained multi-modal models (CLIP and FLAVA), we encode both visual microstructural data and textual expert assessments into shared representations. To overcome limitations in general-purpose embeddings, we develop a customized similarity-based representation that incorporates both positive and negative references from expert-annotated images and their associated textual descriptions. This allows zero-shot classification of previously unseen microstructures through a net similarity scoring approach. Validation on an additively manufactured metal matrix composite dataset demonstrates the framework's ability to distinguish between acceptable and defective samples across a range of characterization criteria. Comparative analysis reveals that FLAVA model offers higher visual sensitivity, while the CLIP model provides consistent alignment with the textual criteria. Z-score normalization adjusts raw unimodal and cross-modal similarity scores based on their local dataset-driven distributions, enabling more effective alignment and classification in the hybrid vision-language framework. The proposed method enhances traceability and interpretability in qualification pipelines by enabling human-in-the-loop decision-making without task-specific model retraining. By advancing semantic interoperability between raw data and expert knowledge, this work contributes toward scalable and domain-adaptable qualification strategies in engineering informatics.

# 1. Introduction

Additive manufacturing (AM) or 3D printing enables the manufacturing of engineering materials that can be tailored to specific applications [1]. The qualification of such additively processed materials largely depends on manual efforts due to their inherent reliance on downstream data processing and expert knowledge. A notable disconnect exists between the raw data generated from material characterization of the engineered materials and the interpretive expertise necessary for their evaluation. This gap impedes productivity in industrial settings by introducing substantial delays and additional efforts associated with the processing of raw data, the extraction of meaningful information, and the qualification of the extracted information.

Raw data obtained from the characterization of engineered materials are typically processed using explicit rule-based algorithms designed to extract relevant microstructural information [2]. Although these algorithms expedite information extraction, their effectiveness is limited when handling inherent variability and noise in materials data. Statistical methods such as deep learning (DL) based semantic segmentation can address this limitation and enhance the raw visual data by transforming it into quantifiable representations [3, 4]. However, even with improved quantification through semantic segmentation, expert evaluation remains necessary to interpret these results against set characterization criteria (e.g., rules to interpret the extent of phases and anomalies). Manual qualification approaches remain predominant in the industry. Therefore, fully automated and rapid qualification methods are needed to meet the industrial requirements for high-throughput analysis.

Textual descriptions of expert knowledge provide a more effective way to capture the characterization criteria as compared to fixed rule-based methods. The recent advancements in vision-language models (VLMs) present a promising opportunity to integrate the enhanced microstructural information produced by semantic segmentation DL models with textual descriptions of the expert characterization knowledge. Textual descriptions can be encoded to explicitly represent characterization expertise across various criteria and enable the interpretation of meaningful information from inherently variable processed microstructural data. These VLMs, designed for joint understanding across vision and language modalities, have demonstrated significant potential in different engineering disciplines, including mechanical design [5], information extraction [6], agile manufacturing [7], as well as generic computer vision and natural language processing tasks [8]. This capacity to create meaningful cross-modal connections emerges from the latent vision-language representations (VLRs), which are learned during their pre-training on large-scale and general-purpose datasets.

While the capabilities of VLMs hold significant promise and have been successfully applied to various scenarios, their direct application to microstructural characterization presents several challenges. Fully automating this process in the context of advanced manufacturing remains difficult due to key limitations that must first be addressed. Data scarcity remains the primary concern, which hinders the ability to fine-tune large pre-trained models to encode domain-specific characterization knowledge [9, 10]. Furthermore, multi-phase materials often exhibit co-existing phases and associated anomalies that require qualification against multiple competing criteria. This complexity necessitates customized evaluation pipelines capable of managing variability in visual and textual modalities and linking the two modalities. These requirements from both the vision (microstructural patterns) and the language (expert knowledge) modalities constrain the straightforward application of generic multi-modal models. Therefore, it is essential to effectively tailor these models to the domain requirements to leverage the cross-modality understanding.

To address these limitations, this work explores the capability of VLMs to bridge visual representations of microstructural information with expert characterization knowledge using a heterogeneous material system produced via an AM process as a case study. We evaluate the encoded vision and language understanding based on feature representations derived from two pre-trained VLMs. Building on this, we examine the potential of joint vision-language understanding by quantifying the semantic similarity between expert assessment criteria and a range of microstructural representations. The trained representations are enhanced in two different ways (customization and hybridization) to effectively link microstructural information with characterization knowledge. Depending on the suitability of a pre-trained representation to provide compelling vision(input)-vision(information) and vision(input)-language(knowledge) linkage, hybridization between different VLMs is proposed. Customizing extracted VLRs is proposed to enhance the similarity evaluation and embed information across microstructural images and knowledge across characterization criteria.

This study ultimately contributes to advancing the linkage between information-level features and knowledge-level understanding, with the following contributions:

- Provides an information-level dataset of microstructures produced by directed energy deposition AM process
- Investigates vision-language models for linking meaningful microstructure information with expert characterization knowledge
- Proposes customized and hybrid vision-language representations (VLRs) for rapid characterization of advanced materials
- Validates the effectiveness of proposed VLRs on optical metallographs from additive manufacturing
- Implements a multi-modal characterization knowledge base for customized VLRs

The rest of this paper is structured as follows. Section 2 presents related works and highlights the existing challenges in materials characterization and knowledge representation. Section 3 presents the overall framework and introduces different levels of informatics representations across the characterization pipeline. The materials characterization data, information, knowledge, and wisdom (application) pyramid are discussed. The optical metallography dataset, the models, customization of multimodal representations, and associated metrics are discussed in Section 4, which introduces the dataset, the methods to customize multimodal representations, and the associated metrics. Section 5 covers the discussions on cross-modality understanding for materials characterization data and presents qualification results derived from customized and hybrid VLRs. Section 6 introduces a multi-modal industrial characterization knowledge base as a potential application of the proposed methodology. Section 7 concludes the paper by listing key contributions and future works.

## 2. Related Works

Engineering informatics is evolving due to advances in data science and its overlap with classical information representation techniques for applications in design, manufacturing, and material science domains [11]. Within this landscape, efforts to digitally represent and analyze microstructural phenomena have evolved from rule-based image processing to sophisticated learning-based frameworks. However, existing systems treat images, measurements, and textual descriptions as disjoint representations. Therefore, the existing workflows are missing the opportunity to fuse them into coherent and retrievable representations [12]. Recent advances in multi-modal representation learning, particularly through vision and language models, have opened new pathways for linking visual and semantic information in engineering datasets [13, 14]. Coupled with developments in materials knowledge graphs [15], digital twins [16], and qualification frameworks [17], these innovations point toward a future in which microstructural data can be semantically searched, interpreted, and reused at scale. In this section, we review key contributions across vision-based microstructure analysis, multi-modal learning, structured domain knowledge representation, and data-driven qualification strategies, each of which informs and motivates our proposed framework.

Early attempts to automate microstructural evaluation relied on classical digital image analysis pipelines [4]. Recent surveys show that convolutional and transformer-based DL models now dominate, providing state-of-the-art accuracy in grain-size estimation [18], phase segmentation [3], and defect detection across alloys and additively-manufactured metals [19]. Transformer architectures and their derivatives have proven especially effective at capturing long-range morphology, motivating a growing body of work on super-resolution and stochastic reconstruction of representative volume elements from limited experimental data [20]. Nevertheless, these purely visual approaches cannot leverage the rich textual metadata (e.g., process parameters, test results, standards, expert criteria) usually required alongside micrographs during industrial qualification.

Efforts in materials knowledge representation have produced large, automatically extracted graphs that link materials, processes, properties, and characterization methods mined from the scientific literature [15]. Domain-specific graphs have also been proposed for finer-grained structural descriptors [21]. A recent contribution by Ye et al. [22] presents the Materials Knowledge Graph (MKG), which leverages large language models (LLMs) to extract and organize materials science knowledge from literature into a structured, ontology-based graph. Studies have also combined knowledge graphs with language models [23]. While these resources encode valuable expert knowledge, they are predominantly text-centric: the microstructure itself is represented only through derived descriptors or embeddings, not through direct linkage with the raw image representations. Consequently, these graphs fall short of capturing variations in the vision domain and linking them with the inherent variations in the text domain.

In parallel with the knowledge-based advancements, the emergence of foundation models, particularly LLMs and vision transformers (ViTs), has begun to reshape the landscape of engineering informatics [24]. LLMs have demonstrated the capacity to extract and reason over complex scientific knowledge related to microstructure evolution [25], defect classification [26], and property prediction [27], often without requiring domain-specific re-training. These models increasingly support tasks, such as literature mining, intelligent process optimization, and the automated interpretation of structure-property relationships. On the visual side, ViT-based architectures have shown superior robustness and generalizability compared to traditional convolutional counterparts [3], especially in low-data regimes common in materials science research. Recent developments in visual foundation models and vision-language transformers have extended this capability by enabling few-shot and zero-shot interpretation of microscopy images across scales and modalities [28]. Together, these trends suggest a convergence toward foundation model-driven materials informatics where pre-trained representations of text and image modalities are leveraged for unified, flexible, and scalable microstructural analysis. As a result, their integration into qualification workflows can enable semantic traceability and process-specific relevance [29-31].

Inspired by the success of models for cross-modality linkage, several works have begun to couple large-scale vision encoders with language models so that visual features and textual descriptions share a common embedding space [32]. Engineering applications of these models include design [33], manufacturing [34], construction [35], material science [36], and more. The Cephalo family of multi-modal vision-LLMs, for example, integrates visual encoders with autoregressive text decoders and is fine-tuned on thousands of figures and captions from materials science papers. It can answer free-form questions about complex fracture patterns and even generate bio-inspired designs from example photographs [36]. On a broader scale, perspectives on the impact of LLMs on materials informatics argue that such multi-modal representations will soon enable automated annotation, query, and generation of 3-D microstructure datasets [37, 38]. To date, however, demonstrations remain at the laboratory-scale as they rarely integrate domain-specific vocabularies or satisfy the traceability requirements of industrial standards.

Existing works do not enhance generalized vision-language embeddings capable of unifying diverse imaging modalities (e.g., micrographs, X-ray computed tomography (XCT) slices, or surface profilometry) with associated textual inspection reports and expert qualification criteria. Across industries, such as aerospace and biomedical manufacturing, the cost and time associated with qualifying new alloys or printing parameters are primarily driven by the reliance on post-process evaluation protocols. To mitigate this bottleneck, recent research has explored hybrid qualification approaches that integrate microstructure-based simulations with a limited set of mechanical experiments to accelerate acceptance decisions without compromising reliability [39]. Complementary efforts have introduced digital twin frameworks that embed real-time sensor data into physics-informed models to facilitate in-situ quality assurance, particularly in AM-based systems [40]. While these studies highlight the economic and operational value of linking process conditions to structural metrics, the computer vision components involved are typically domain-specific classifiers or segmentation networks tailored to individual tasks.

The literature thus reveals three disconnected themes: (1) high-capacity computer vision models that excel at recognizing morphology, but ignore characterization semantics; (2) emergent multi-modality models capable of cross-modal reasoning, but not yet tuned to engineering taxonomies or qualification protocols; and (3) knowledge graphs and digital twins that capture process–structure–property linkages, but lack domain grounded qualification. Several manufacturing applications of engineered materials can benefit from data-driven rapid qualification enabled by the linkage between vision models elevating raw data to meaningful information and emerging VLMs capable of linking this information with expert characterization knowledge.

By customizing VLM representations on paired micrographs of a heterogeneous material system and associated qualification assessments, our study aims to bridge these themes. It also provides opportunities for multi-modal industrial qualification knowledgebases grounded in enhanced VLRs. The result is a unified framework that supports content-based retrieval (e.g., find similar microstructures), semantic search (e.g., find micrographs with specific properties), and quantitative comparison (e.g., multi-modal similarity) across samples, thus reducing the time and data burden required for industrial qualification. Such an approach advances applied engineering informatics by demonstrating how multi-modal representation learning, domain understanding, and manufacturing-oriented knowledge can be effectively combined to deliver actionable insights to engineers and operators on the shop floor.

## 3. Overall Framework

The proposed framework follows the well-known Data–Information–Knowledge–Wisdom (DIKW) hierarchy, where different semantic levels in the materials characterization pipeline are arranged accordingly, as shown in Figure 1. At the framework's base lies data, representing unstructured material features in pixel intensities. Once the raw microstructural features are segmented and grouped based on shared characteristics, such as intensity thresholds or geometric contours, they transition into information. The raw data used in this work represent microstructural images generated from optical metallography performed on a nickel tungsten carbide (Ni-WC) metal matrix composite (MMC). In the context of Ni-WC MMC microstructures, this stage reveals physical features, such as carbide clusters, porosity distributions, heat-affected regions, and anomalies related to the processing conditions. These informational patterns evolve into knowledge as domain-specific interpretation is applied, such as correlating carbide morphology with mechanical performance or thermal gradients. This level enables experts to determine material behavior and processing implications supported by multi-scale feature representations. Wisdom is at the top of the hierarchy, where application-specific rules, such as acceptable ranges of defect severity or carbide dispersion needed for application domains (e.g., mining, agriculture, aerospace), are applied to support qualification decisions. The annotated pyramid structure highlights this semantic progression using representative microstructural examples, emphasizing how data-driven insights can be transformed into contextualized expert guidance within materials informatics workflows.

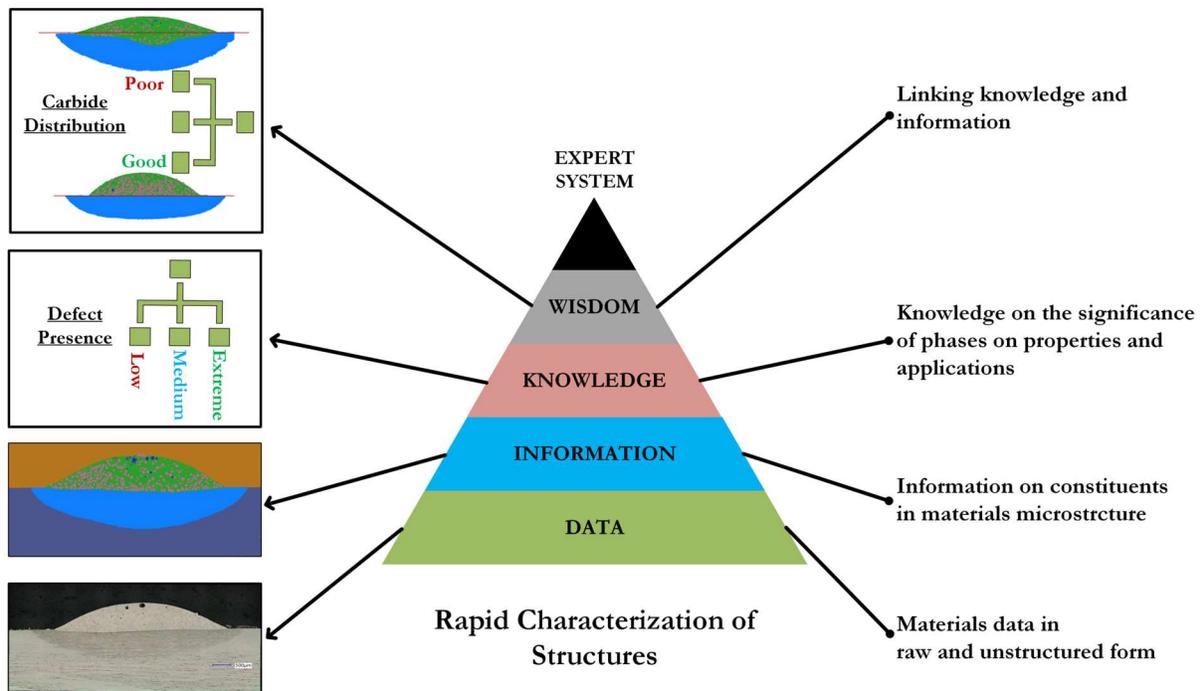

Figure 1: DIKW hierarchy for rapid qualification of engineered materials in industry

Figure 2 presents a layered view of the material characterization process following the semantic levels. The framework begins with the deposition of structures via the directed energy deposition AM process [41]. Raw data characterizing the process and resulting samples are generated through subsequent evaluation steps, such as optical metallography. These raw micrographs constitute the data level, capturing intensity-based representations of microstructural constituents. The first informatics interface, the data-to-information transition, occurs through semantic segmentation techniques, where DL models extract structural features (e.g., phases, boundaries, defects). These features are then translated into information, such as area fractions, size distributions, and defect presence, which summarize relevant microstructural characteristics.

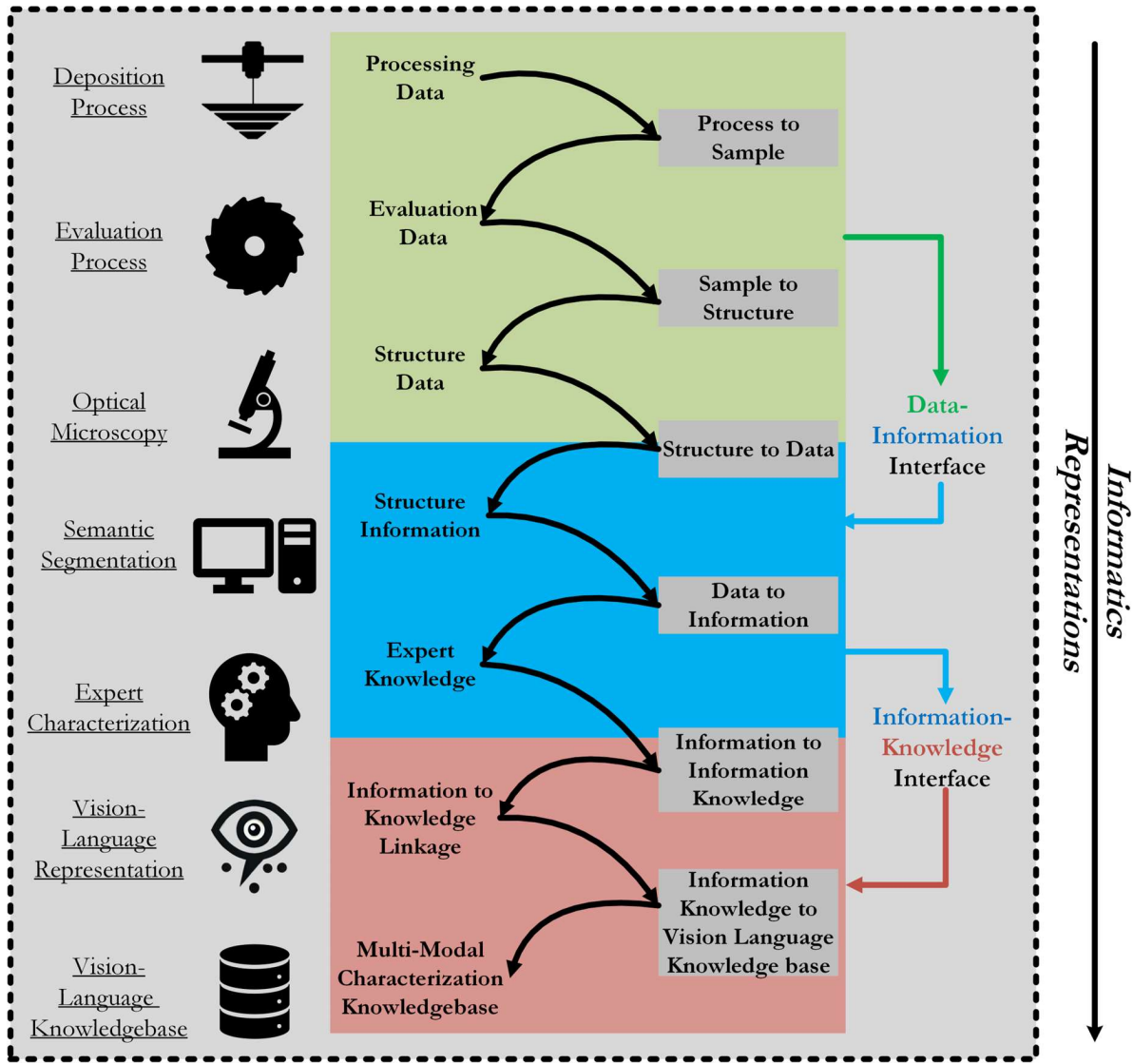

Figure 2: Flow of informatics representations across the characterization pipelines of advanced manufacturing processes

The second informatics interface bridges the information and knowledge levels, where extracted microstructural statistics are interpreted using expert-defined criteria, decision thresholds, or rule-based systems. This expert knowledge is often task-specific, but lacks formal semantic encoding, limiting its reuse and scalability. The final contribution of this framework lies in the wisdom layer, where a new representation interface based on vision-language models is introduced. Here, vision encodes microstructural features, while language encodes expert knowledge, forming a multimodal characterization knowledge base. This cross-modality linkage allows for a more integrated, application-aware reasoning system that aligns with industrial qualification needs by semantically linking image-derived information with domain knowledge in a reusable and extensible format.

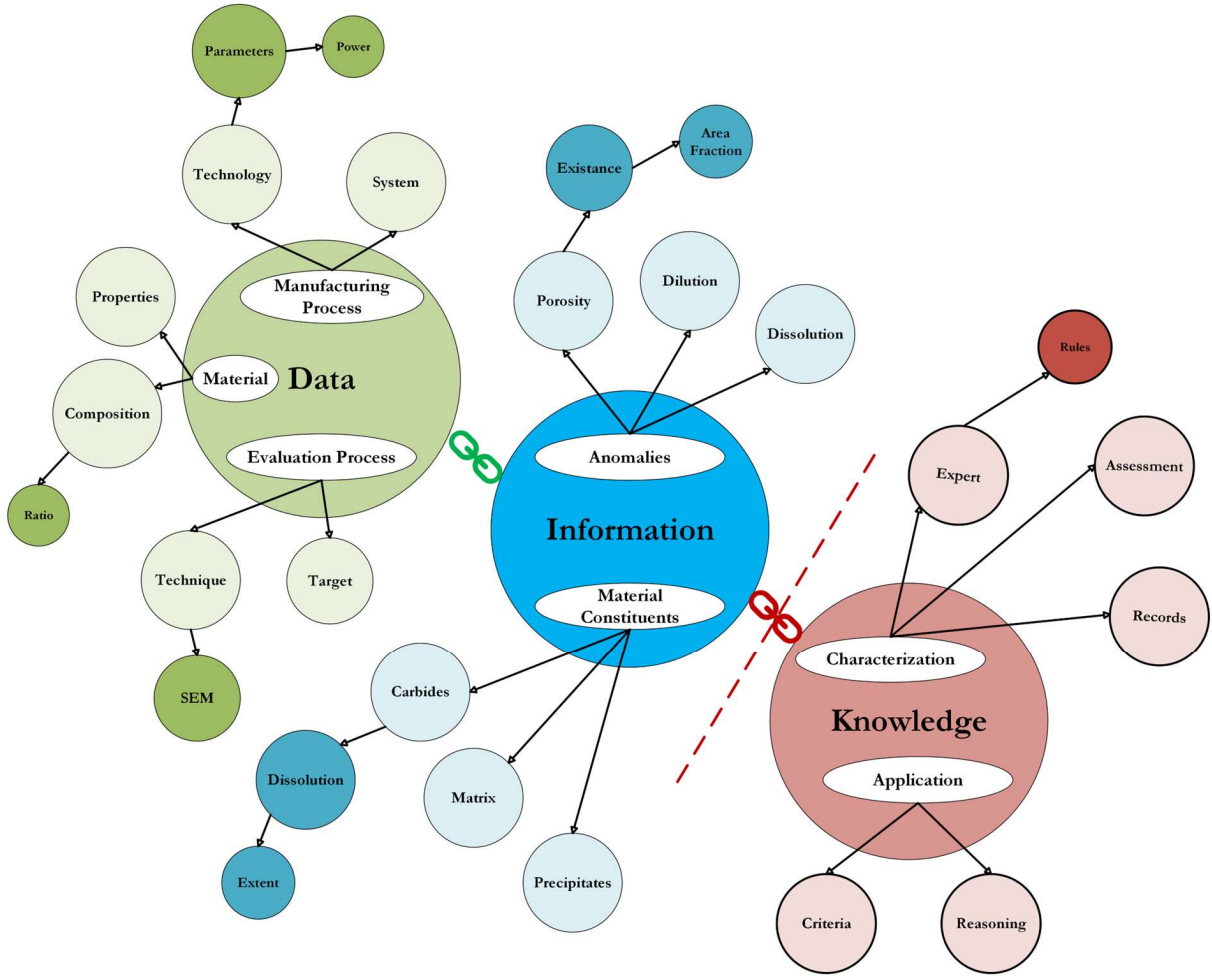

Figure 3: Hypothetical representation of data, information, and knowledge modules. The data-information (green link) is well established in materials characterization whereas the information-knowledge (red link) is mostly manual and lacks automation in industry

In the context of materials characterization for advanced manufacturing, semantic levels, such as data, information, and knowledge represent stages in the transformation of raw experimental observations into engineering decisions. At the data level, we handle inputs like processing parameters or microscopy images; these are elevated to information through methods such as DL-based semantic segmentation, which naturally establish links by extracting features like phases, defects, or geometries. At the knowledge level, this information is interpreted using expert judgment to qualify materials or predict performance. While data-to-information connections are increasingly enabled by DL models trained on annotated micrographs, the information-to-knowledge interface remains largely manual and domain-dependent. As illustrated in Figure 3, although conceptual inter- and intra-level linkages exist, current systems lack generalizable frameworks for formalizing expert reasoning. Here, VLMs offer a promising solution, inherently capable of bridging information and knowledge levels, by embedding both visual features and textual characterizations into a semantically shared space that can generalize across applications and materials systems.

## 4. Methodology

This section introduces the experimental dataset, alongside the associated microstructure information and characterization knowledge. The models used to extract pre-trained VLRs are discussed. The customization of extracted embeddings to enhance the information-knowledge linkage is presented.

A) Metallography Dataset

The dataset represents a heterogeneous material system from MMCs containing a metal matrix and reinforcement particles manufactured through an AM process. The complexity of the material system and its dependence on expert knowledge for quality evaluation are representative of a typical engineered material that can consist of multi-phase constituents. Therefore, selecting Ni-WC MMC for the case study involving information and knowledge linkage can provide insights for the application of proposed framework. The data was generated through an AM process, specifically powder-fed laser directed energy deposition [41]. The primary source of the microstructure data was an optical metallography procedure, which revealed the material microstructural data for downstream processing. Metallographic analysis of the single samples consisted of sectioning the additively deposited beads transverse to the travel/deposition direction at the center of the bead. These metallographic cross-sections were mounted in Bakelite and manually ground successively with finer resin-bonded diamond abrasive pads (80 grit, 180 grit, 220 grit, 500 grit) and water to produce a flat ground surface using a Struers LaboPol-2 grinding and polishing machine. After grinding, the metallographic samples were polished to a mirror finish using the same Struers machine and an affixed LaboDoser for automated dosing of progressively finer diamond slurries (9 µm, 6 µm, 3 µm) as well as a 0.04 µm alumina suspension for the finishing step. The heat-affected zone (HAZ), resulting in the substrate from each deposited bead, was revealed by immersing for 5 s in a Nital etch solution (3 ml nitric acid in ethanol) followed by rinsing and cleaning with high-purity ethanol. After preparation, the metallographic samples were examined under a Keyence VHX-7000 microscope at 50X magnification. The Keyence software was then used to automatically stitch the images together, creating a large panorama of the whole cross-section with a 0.20 µm/pixel resolution. A typical metallographic cross-section with informatics levels is shown in Figure 4.

B) Microstructure Information

The extracted images were further post-processed in Adobe Photoshop to improve image quality and consistency. The procedure was as follows: straightening the image relative to the fusion line between the bead and the base material, cropping to an appropriate size to include bead reinforcement area alongside the substrate, masking the Bakelite using a black background, increasing the contrast to improve the visual difference between the tungsten carbides and the surrounding nickel-based metal matrix, applying an "Unsharp Mask" filter, and placing the scale bar in the corner of the image. The processed images were then segmented using a DL-based semantic segmentation tool, MicroSegQ+, developed for quantitative metallography [3]. The results provided useful structure information for applying domain-based characterization knowledge. The evaluation sequence was repeated for all the additive samples manufactured during the experiments. A sample bead cross-section depicting a panoramic view after stitching individual cross-sectional images taken from the optical microscopy dataset is shown in Figure 4a. A segmented version of the same metallograph is presented in Figure 4b and highlights the grouping of specific intensity ranges that represent constituents of interest in the microstructure. These phases include a red fusion line, a blue HAZ, a green bead reinforcement area, pink carbide particles, and dark blue porosity pixels. The presence or absence of these constituents across the metallograph is then interpreted using domain knowledge, as highlighted in Figure 4c. Together, these informatics' levels help qualify the newly generated microstructural images.

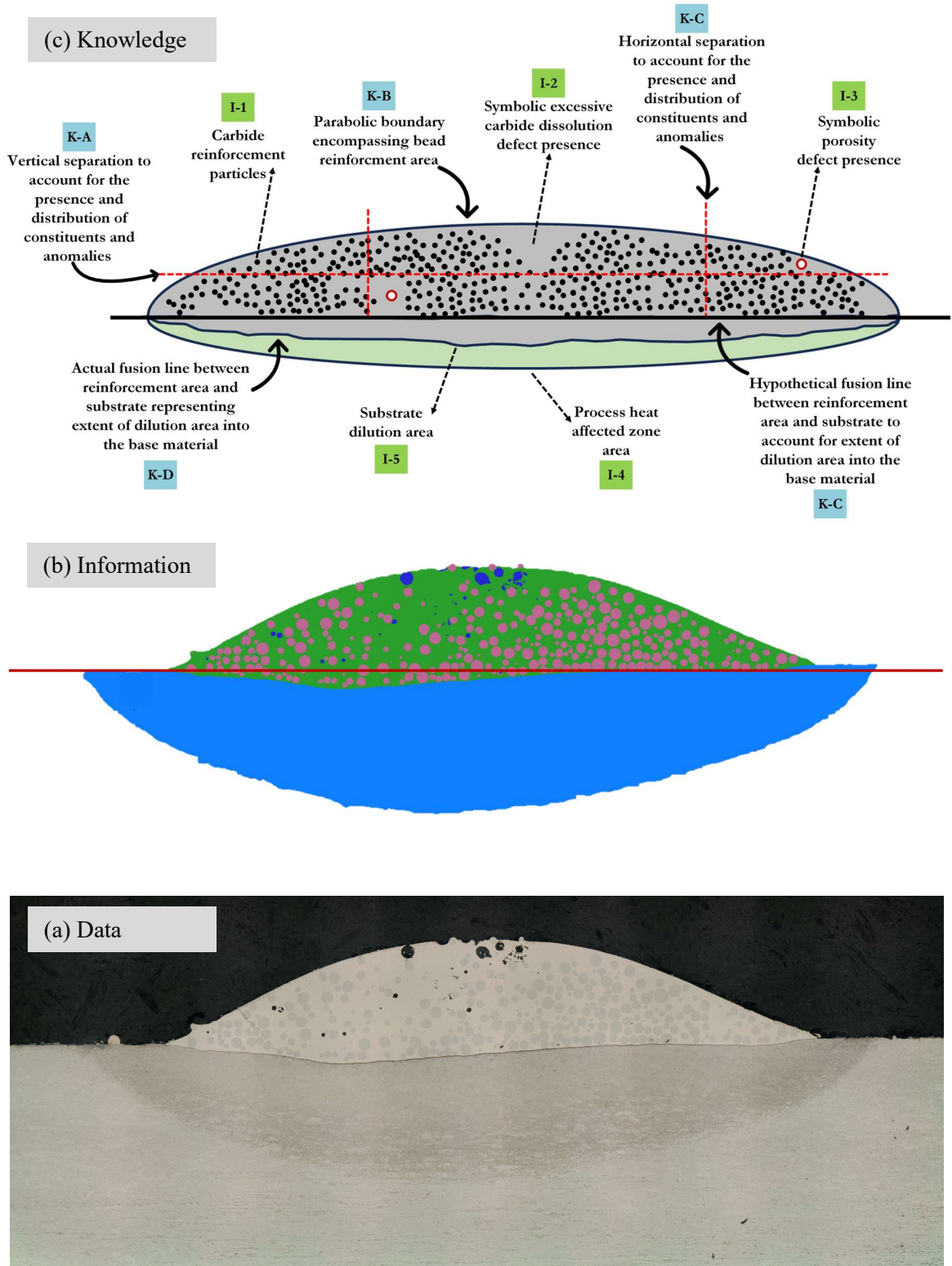

Figure 4: Data (a/bottom), information (b/middle) and knowledge (c/top) representations of bead reinforcement area. Application knowledge is constant and known whereas the physical data and meaningful information is a variable

C) Characterization Knowledge

The knowledge level of the framework involves a sequence of expert assessments (EA) designed to evaluate the quality and integrity of semantically segmented cross-sections obtained from optical metallographs of Ni-WC MMCs. This process transforms segmented image data (information) into qualified samples (e.g., accepted or rejected) by applying domain-specific qualitative and quantitative criteria grounded in materials science and manufacturing practice. This characterization process for Ni-WC MMCs can be represented by the following six expert assessments (EA-1 to EA-6), which are also structured in Figure 5:

- **Dilution Assessment (EA-1):** This assessment quantifies the extent of dilution of the nickel-based matrix into the steel substrate. Following established laser cladding literature, dilution is typically characterized using three criteria: maximum depth at the center, dilution area, and elemental migration (via chemical mapping). The most practical and widely used metric is the percentage area of dilution, with 1–2% considered acceptable and 5% regarded as excessive. Experts often assess this visually by drawing a straight horizontal line from the steel substrate and evaluating the area below the fusion line.
- **HAZ Assessment (EA-2):** This step evaluates the thermal impact on the substrate, particularly the depth of the HAZ beneath the deposited bead. A straight vertical line is drawn from the center of the bead to the substrate to measure HAZ depth. While some practitioners consider area-based metrics, they are often unreliable due to bead size variations. A depth threshold of ≤1 mm is typically used in practice, though alternative criteria, such as HAZ hardness, may be specified depending on material systems and customer requirements.
- **Bead Reinforcement Area Validation (EA-3):** This assessment determines whether the deposited bead provides a distinct reinforcement area above the substrate. Cross-sections that fail to show a defined bead structure are immediately disqualified from further evaluation and are categorized as failed due to the lack of fusion or complete detachment from the substrate.
- **Porosity Assessment (EA-4):** Porosity is examined to distinguish between acceptable and unacceptable defects. Minor porosity resulting from the fallout of carbide particles during the preparation process is generally ignored. However, significant porosity from gas entrapment or solidification shrinkage is representative of excessive process heat and typically necessitates process modification. Experts focus on pore size, area fraction, and spatial distribution with particular attention to features that disrupt bead integrity.
- **Carbide Dissolution Assessment (EA-5):** This assessment identifies signs of partial or complete dissolution of tungsten carbide particles within the reinforcement area. The presence of jagged or reprecipitated carbides indicates thermal degradation in the raw metallography data. A threshold of approximately 30% carbide area fraction is often used as a lower bound for acceptable reinforcement quality, although this value may vary depending on specific application requirements.
- **Carbide Distribution Assessment (EA-6):** Uniform distribution of carbide particles is critical for ensuring consistent wear resistance across the cross-section. Experts evaluate both vertical and horizontal distribution. Vertically, inconsistencies between the central and toe regions of the bead may lead to premature wear. Horizontally, significant variation in carbide concentration from top to bottom may signal poor melt pool dynamics or flotation effects. These defects are especially relevant in multi-layer deposits, where heat accumulation can exacerbate carbide redistribution. Only cross-sections that satisfy the criteria established in all six expert assessments are classified as acceptable. Cross-sections exhibiting excessive porosity, dilution, carbide dissolution, or non-uniform distribution are rejected. This structured and expert-guided process ensures that only high-integrity sections proceed for further analysis or application.

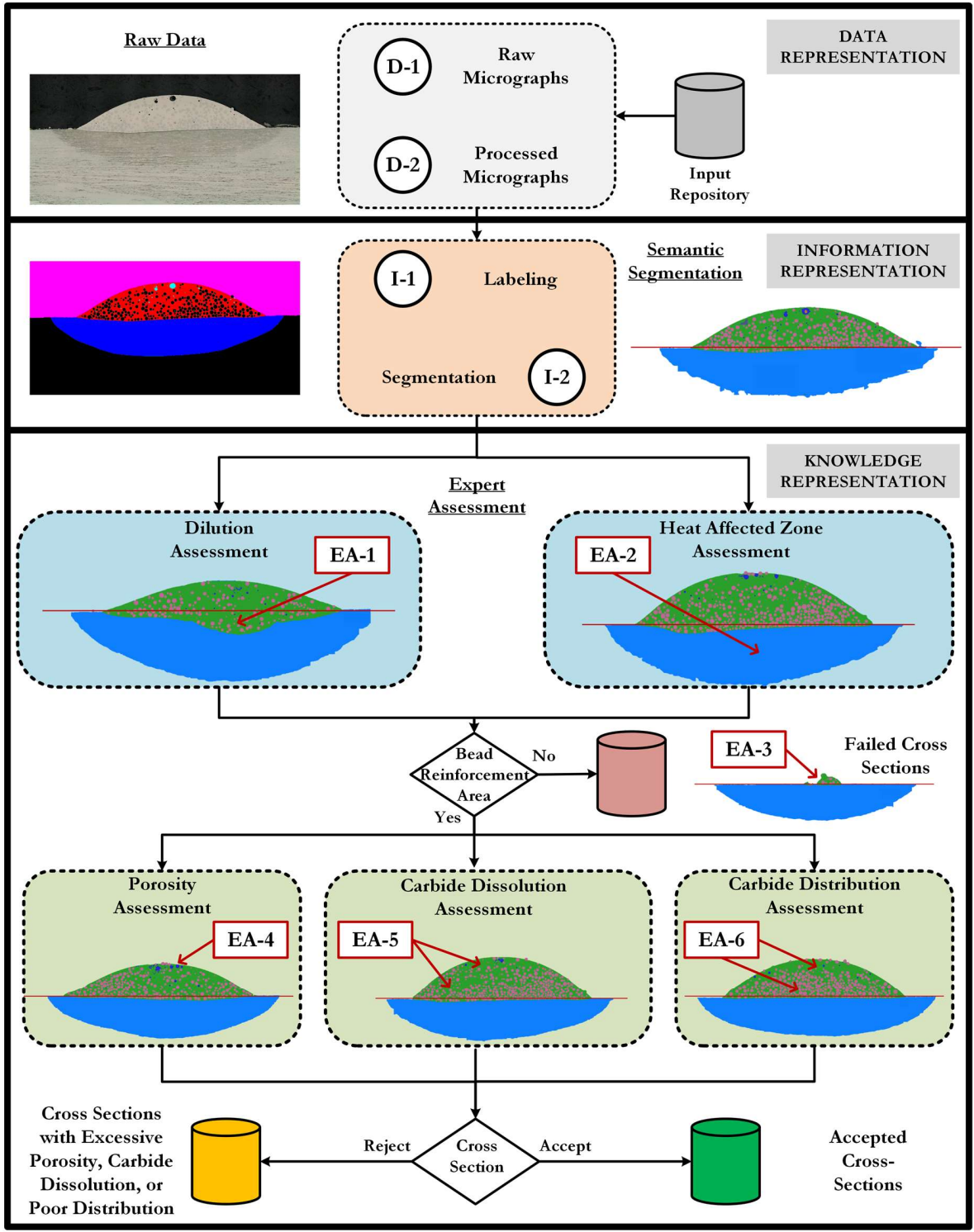

Figure 5: Qualification sequence for additively manufactured Ni-WC MMC microstructures highlighting different expert assessment criteria

The aforementioned expert assessments have been encoded as textual representations to serve as natural language descriptions of key structural attributes within segmented metallographs (see Table 1). These qualitative descriptions act as domain-grounded rules, capturing the visual criteria materials scientists use to evaluate microstructural integrity, such as dilution, porosity, and reinforcement distribution. Unlike categorical annotations, these text-based representations provide semantically rich, compositional information that aligns more naturally with the pre-training objectives of VLMs (e.g., CLIP, FLAVA models). The goal of encoding such representations is to enable image-text linkage through similarity metrics in a shared multi-modal embedding space, thereby facilitating zero-shot classification, image retrieval, and interpretability without requiring task-specific retraining. Each positive and negative representation forms a structured criterion that guides the semantic reordering of images based on expert judgment. While the quantitative representation of microstructure can only be provided by the pixel-level visual patterns, the complementary textual representations encode high-level structural knowledge, making them suitable for cross-modal reasoning. This framework bridges domain expertise and machine reasoning by grounding abstract visual patterns in explainable texts, which enhances the applicability of VLMs for scientific qualification tasks.

Table 1: Example color-aware textual representation of the expert assessments performed on segmented metallographs

| Assessment Component | Text Descriptions |
|---|---|
| Expert Assessment 1 (Dilution) | Positive Representation: "An ideal microstructural image has the bead reinforcement area fully contained above the red fusion line, highlighting no dilution of the base/substrate material." <br> Negative Representation: "A non-ideal microstructural image shows the bead reinforcement area intruding below the red fusion line, indicating dilution of the base/substrate material." |
| Expert Assessment 2 (Heat Affected Zone) | Positive Representation: "An ideal microstructural image has a small HAZ in blue under the red fusion line, representing the structurally altered portion of the base/substrate." <br> Negative Representation: "A non-ideal microstructural image shows an excessively large blue HAZ under the red fusion line, suggesting excessive thermal damage to the base/substrate." |
| Expert Assessment 3 (Reinforcement Area) | Positive Representation: "A green bead reinforcement area with pink carbide particles is present above the blue substrate."" <br> Negative Representation: "A non-ideal microstructural image shows an incomplete or poorly defined green bead reinforcement area with missing or sparse pink carbide particles above the base substrate." |
| Expert Assessment 4 (Porosity) | Positive Representation: "An ideal microstructural image is free from porosities or voids (highlighted as dark blue pixels) in the bead reinforcement area." <br> Negative Representation: "A non-ideal microstructural image contains visible porosities or voids (dark blue pixels) scattered throughout the bead reinforcement area, indicating lack of fusion or gas entrapment." |
| Expert Assessment 5 (Dissolution) | Positive Representation: "An ideal microstructural image has minimal dissolution of the pink carbide particles, resulting in a reinforcement area visibly covered with the pink carbide particles." <br> Negative Representation: "A non-ideal microstructural image shows severe dissolution of the pink carbide particles, resulting in a reinforcement area lacking visible carbide coverage." |

| Expert Assessment 6 (Distribution) | Positive Representation: "An ideal microstructural image has uniformly distributed pink carbide particles in the green metal matrix of the bead reinforcement area above the fusion line." <br> Negative Representation: "A non-ideal microstructural image has uneven or clustered pink carbide particles in the green metal matrix above the fusion line, indicating poor reinforcement distribution." |
|---|---|

### D) Vision and Language Representations

VLMs have emerged as a transformative class of multi-modal learning frameworks that enable joint reasoning over visual and textual inputs. At a high level, these models can analyze a new, unseen image by comparing it to a large set of concepts learned from image-text pairs, effectively matching the image to the most semantically relevant textual descriptions. For instance, a VLM trained on annotated image-text pairs can interpret a new microstructural image by identifying closely aligned expert criteria in the textual representation. These models are typically pre-trained on large-scale corpora of image-text pairs, learning aligned representations that capture semantic associations across modalities. VLMs enable applications like image captioning, visual question answering, cross-modal retrieval, and zero-shot classification by learning to represent visual and textual information in a common semantic space. In technical fields such as materials informatics and manufacturing, they offer new opportunities to incorporate expert knowledge into visual inspection processes. Here, expert annotations or domain standards can be expressed as natural language prompts and evaluated against visual evidence from microstructural images, enabling semantically grounded inference in high-dimensional, label-scarce settings.

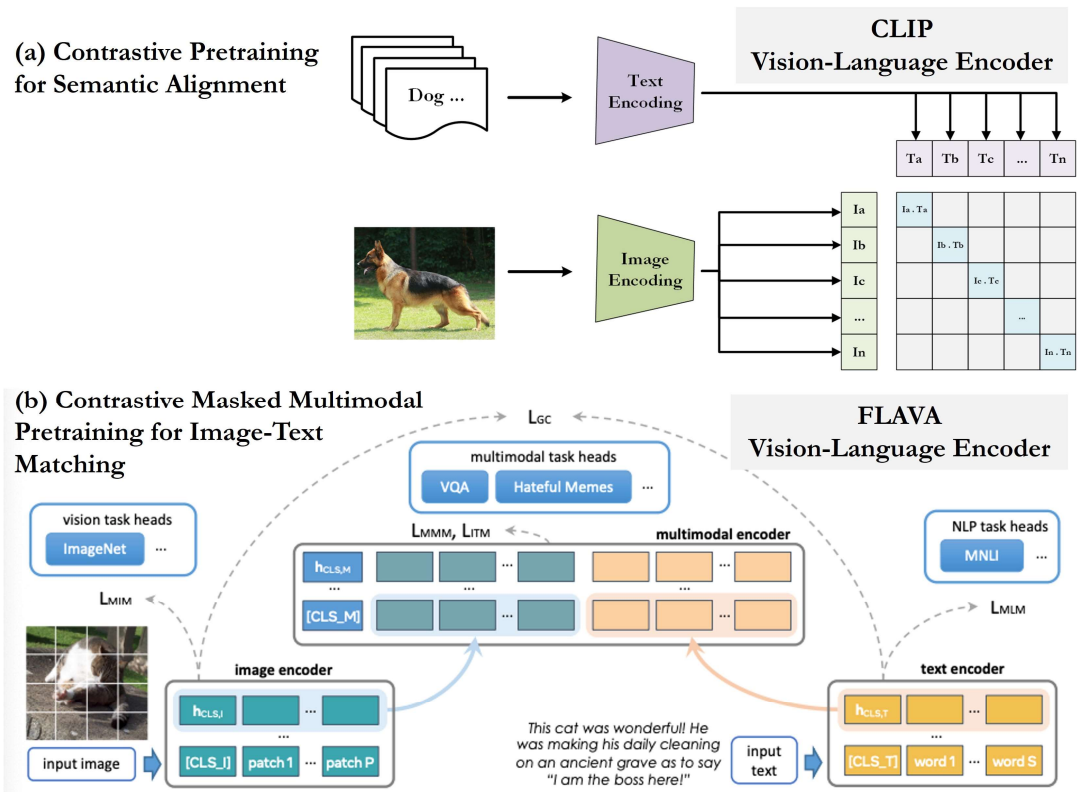

Figure 6: CLIP and FLAVA encoders during pre-training. Figures adapted from [42] and [43] with symbols defined in the text

Among the most widely adopted and representative VLMs are CLIP (Contrastive Language-Image Pre-training), developed by OpenAI [42], and FLAVA (Foundational Language and Vision Alignment), developed by Meta AI [43]. These models exemplify two distinct design philosophies for learning visual-textual alignment. CLIP employs separate image and text encoders trained via contrastive learning, where the goal is to align embeddings of matching image-text pairs while separating mismatched ones. This results in a shared embedding space optimized for cross-modal similarity, supporting zero-shot classification by comparing an image's embedding to textual prompt embeddings using cosine similarity. FLAVA, by contrast, introduces a unified architecture that combines unimodal image and text encoders with a multi-modal fusion encoder. It is trained using a hybrid objective that includes contrastive loss, masked language and image modeling, and image-text matching. This enables FLAVA to support both unimodal understanding and multi-modal reasoning tasks within a single transformer-based framework. The multi-modal encoders of CLIP and FLAVA used in the pre-training are adapted from the sources and presented in Figure 6 for reference. CLIP has a shared image-text embedding space, whereas the FLAVA model uses unimodal and multi-modal encoder to learn the image-text linkages.

The CLIP loss is presented by Equation 1 below:

$$clip\ loss = \frac{1}{2N} \sum_{i=1}^{N} [-\log \frac{e^{z_i^I \cdot z_i^T / t}}{\sum_{j=i}^{N} e^{z_i^I \cdot z_i^T / t}} - \log \frac{e^{z_i^T \cdot z_i^I / t}}{\sum_{j=i}^{N} e^{z_i^T \cdot z_i^I / t}}] \qquad (1)$$

Where:

- $f_\theta(x_i)$: Image encoder function parameterized by θ, producing a d-dimensional embedding $z_i^I$ for image $x_i$

- $g_\varphi(t_i)$: Text encoder function parameterized by φ, producing a d-dimensional embedding $z_i^T$ for text $t_i$

- $z_i^I \cdot z_i^T$: Cosine similarity between normalized image and text embeddings (dot product after L2 normalization)

- N: Number of image–text pairs in a batch

- t: Temperature parameter that scales the logits before applying softmax (controls sharpness of distribution)

- $\sum_{j=i}^{N} e^{z_i^I \cdot z_i^T / t}$: Denominator of the softmax, summing over all text embeddings to compute probability distribution for contrastive matching

- $\sum_{j=i}^{N} e^{z_i^T \cdot z_i^I / t}$: Similar softmax denominator for the symmetric text-to-image contrast

- clip loss: Final symmetric contrastive loss, averaged over both image-to-text and text-to-image directions

CLIP provides vision and language features in the same dimension (e.g., 512) but keeps them separate in the semantically aligned space. On the other hand, FLAVA provides vision and language features in different dimensions, while also providing unified multi-modal vision-language features. FLAVA also has similar image-text contrastive loss as CLIP, as shown in Equation 2. The overall contrastive loss is computed symmetrically over both image-to-text and text-to-image directions; this encourages alignment of corresponding pairs, while penalizing mismatches across the batch.

$$contrastive\ loss = -\log \frac{e^{z_i^I \cdot z_i^T / t}}{\sum_{j=i}^{N} e^{z_i^I \cdot z_i^T / t}} \qquad (2)$$

In the FLAVA architecture, the unimodal image and text embeddings, denoted by $h_i^I$ and $h_i^T$, are passed through a cross-modal transformer to produce a fused multimodal representation $h_i^M$, from which the $[CLS_M]$ token is extracted to capture joint semantic information. This fused token is used in a binary classification setup to determine whether a given image-text pair is aligned, yielding the following image-text matching (ITM) loss:

$$image-text\ matching\ loss = -y_i \log \hat{y}_i - (1 - y_i)\log(1 - \hat{y}_i) \qquad (3)$$

The complete FLAVA loss is therefore composed of contrastive loss ($L_{contrast}$), image-text matching loss ($L_{ITM}$), masked image modeling loss ($L_{MMM}$), masked language modeling loss ($L_{MIM}$), and the masked multimodal loss ($L_{MLM}$). Equation 4 represents the weighted version of the total loss.

$$flava\ loss = \lambda^1 L_{contrast} + \lambda^2 L_{ITM} + \lambda^3 L_{MMM} + \lambda^4 L_{MIM} + \lambda^5 L_{MLM} \qquad (4)$$

A key difference between these models is how they construct and utilize joint image-text representations. CLIP performs alignment solely through its independent encoders and relies exclusively on contrastive objectives during pretraining. The resulting embeddings are semantically aligned but not fused, making them ideal for retrieval-style tasks, where inference involves computing cosine similarity between an image embedding and a set of text embeddings. By contrast, FLAVA generates a fused [CLS_M] multi-modal token using cross-attention between image and text features in a shared encoder. This token captures bidirectional contextual dependencies and is suited for downstream reasoning tasks, such as multi-modal classification or visual entailment. However, FLAVA's contrastive alignment, used for retrieval and similarity ranking, relies not on the fused embedding but on projection heads applied to the unimodal encoders' [CLS_I] (image) and [CLS_T] (text) outputs. As such, FLAVA separates contrastive matching from joint semantic reasoning within its architecture, offering broader flexibility across tasks.

To operationalize zero-shot classification (Figure 7) or retrieval using these models, the similarity between an image and a set of candidate text prompts is typically quantified using cosine similarity between L2-normalized embeddings. In CLIP, this similarity defines the model's core inference mechanism, yielding dot-product logits that are often normalized via a softmax function to form a probability distribution over prompts. FLAVA supports a similar mechanism through its contrastive_logits_per_image output, which reflects the alignment between unimodal embeddings. Therefore, FLAVA uses its multimodal [CLS_M] embedding for tasks that require deeper joint image-text reasoning rather than simple similarity comparisons between the two modalities. The similarity scores can rank candidate prompts or labels in both cases, enabling Top-K retrieval, flexible categorization, and interpretable decision-making. This paradigm is particularly well-suited for materials science applications, where domain knowledge can be encoded textually (e.g., "porous microstructure" or "minimal dilution") and evaluated against complex visual data without requiring task-specific retraining.

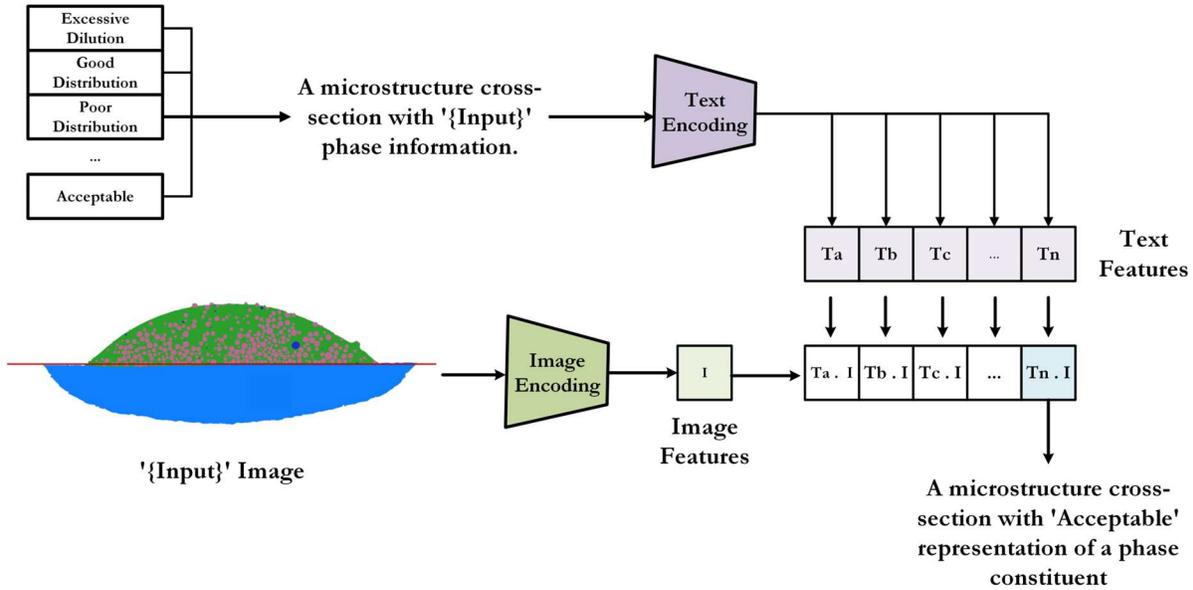

Figure 7: Zero-shot classification with pre-trained multi-modal CLIP and FLAVA representations

E) Customized and Hybrid Representations

A customized strategy was employed to represent expert knowledge in the product qualification process effectively. This approach began by constructing fused feature representations using positive and negative examples drawn from expert-annotated data. Positive examples consist of relevant text prompts and pre-qualified images that reflect acceptable microstructural characteristics, while negative examples capture undesirable features such as excessive dilution or poor distribution. For a given input image, similarity scores are computed separately with respect to the positive and negative sets. Specifically, the similarity to the positive set is obtained by cosine similarity between the input image and the fused positive image set, as well as between the input image and the positive text embeddings. A similar process is followed for the negative set and the net similarity is computed.

The final output score is determined by subtracting the similarity to the negative examples from the similarity to positive ones. This net similarity score is a proxy for how well the input image aligns with the expert-defined criteria for an acceptable microstructure. A non-negative score indicates closer alignment with positive examples and results in a "Positive" label, whereas a negative score suggests greater similarity to negative references and is labeled "Negative." This method enables the application of expert knowledge into an interpretable and adaptable scoring system for microstructural qualification and can be re-applied for a range of qualification criteria.

The decision to use CLIP's embedding for image-text similarity was grounded in several observations. First, the CLIP model performs the image-text semantic alignment in a shared latent space while keeping the two representations separate. This allows for a quick comparison of the incoming image representation with the reference texts. Secondly, using the FLAVA model for the cross-modality linkage would require generating a multi-modal (both image and text) joint embedding for each input image, which will lead to increased computational expense and qualification time. Moreover, extensive experiments would be required to find the most effective textual prompts since the primary input for similarity comparisons will depend on their fusion with the image representations. Similarly, using the FLAVA model for image-to-image similarity was grounded in empirical evidence that its vision encoder provides stronger similarity metrics as it has been pre-trained on unimodal data.

Customizing the pre-trained vision and language embeddings is done through iterative updates. Equations 5-8 formalize the similarity computations using CLIP (vision-language), FLAVA (vision-vision), and their hybridized form with custom positive and negative prompts. The embeddings are L2-normalized, and cosine similarity is used throughout.

1. Customizing vision-language similarity

Given an image embedding $z^I \in \mathbb{R}^d$ and a set of normalized positive and negative text embeddings $z_i^{T+}$ and $z_j^{T-}$, the mean positive and negative text embeddings are computed as follows:

$$\bar{z}^{T+} = (1/N^+) \sum_i z_i^{T+}, \quad \bar{z}^{T-} = (1/N^-) \sum_j z_j^{T-} \quad (5)$$

The CLIP-based similarity delta is then given by:

$$\Delta_{CLIP} = cos(z^I, \bar{z}^{T+}) - cos(z^I, \bar{z}^{T-}) \quad (6)$$

2. Customizing vision-vision similarity

Given a query image embedding $z^\varphi \in \mathbb{R}^d$ and reference image embeddings for positive and negative cases, the average embeddings are:

$$\bar{z}^{I+} = (1/M^+) \sum_k z_k^{I+}, \quad \bar{z}^{I-} = (1/M^-) \sum_l z_l^{I-} \quad (7)$$

The FLAVA-based similarity delta is computed as:

$$\Delta_{FLAVA} = cos(z^\varphi, \bar{z}^{I+}) - cos(z^\varphi, \bar{z}^{I-}) \quad (8)$$

3. Hybridization of embeddings

To combine the CLIP and FLAVA deltas, each is standardized using z-score normalization across all samples to regularize the dynamic range of pretrained modalities within the dataset distribution. Let $\mu_{CLIP}$ and $\sigma_{CLIP}$ denote the mean and standard deviation of $\Delta_{CLIP}$:

$$z_{CLIP} = \frac{(\Delta_{CLIP} - \mu_{CLIP})}{\sigma_{CLIP}}$$

and similarly for FLAVA:

$$z_{FLAVA} = \frac{(\Delta_{FLAVA} - \mu_{FLAVA})}{\sigma_{FLAVA}}$$

The final hybrid similarity score is the sum of the standardized deltas:

$$\Delta_{Hybrid} = z_{CLIP} + z_{FLAVA}$$

The Figure 8 illustrates the sequence of embedding customization and hybridization. Expert assessments and microstructural images for acceptable and failed samples are used to provide reference similarity criteria for the two modalities. Vision and language embeddings from incoming or new images are extracted (1) and compared (2) with reference embeddings. Multi-modal similarity results from different models are fused to generate final scores for labeling the incoming images.

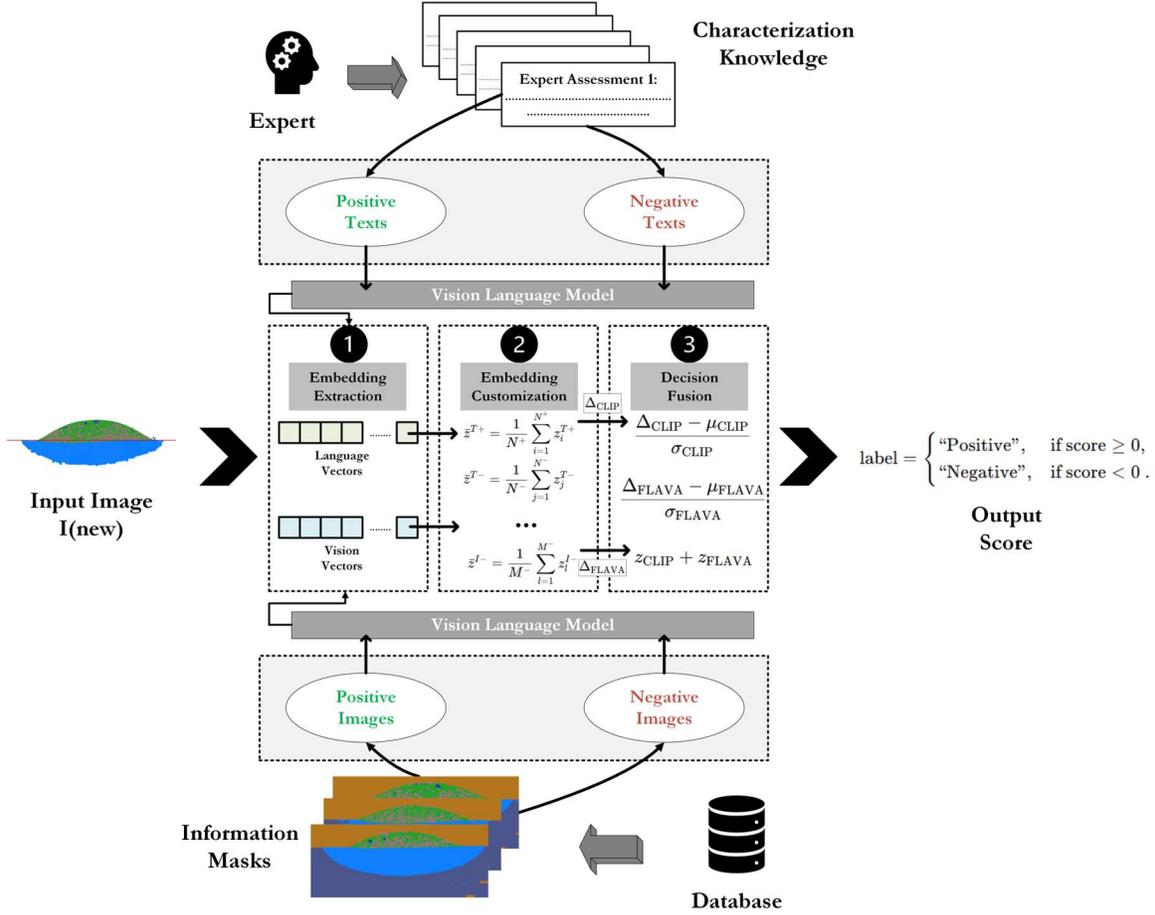

Figure 8: Scalable customization and modular hybridization of pre-trained unimodal and multi-modal representations

### 5) Results and Discussion

The results of expert labeling for the dataset samples used in this work are provided in Table A1 of the appendix. Figure 9 and Figure 10 highlight the vision encoded space of the VLMs that were considered. The results used t-distributed stochastic neighborhood embeddings (t-SNE) to plot 2D representation of encoded features. Although the reduced 2D space based on t-SNE has no physical measure, the relativity of the original high-dimensional space among the inputs is maintained. This highlights inherent similarities and dissimilarities among vision representations of different microstructural images. In both FLAVA and CLIP models, three regions stand out. One such region corresponds to higher bead height, which is highlighted in peach shade. We also see the carbide distribution similarity among the cases with larger cross-sections. For instance, in the peach regions (Figure 9a, Figure 10a) representing higher bead reinforcement areas without dilution, we can see that these reinforcement areas are placed closer together for both CLIP and FLAVA.

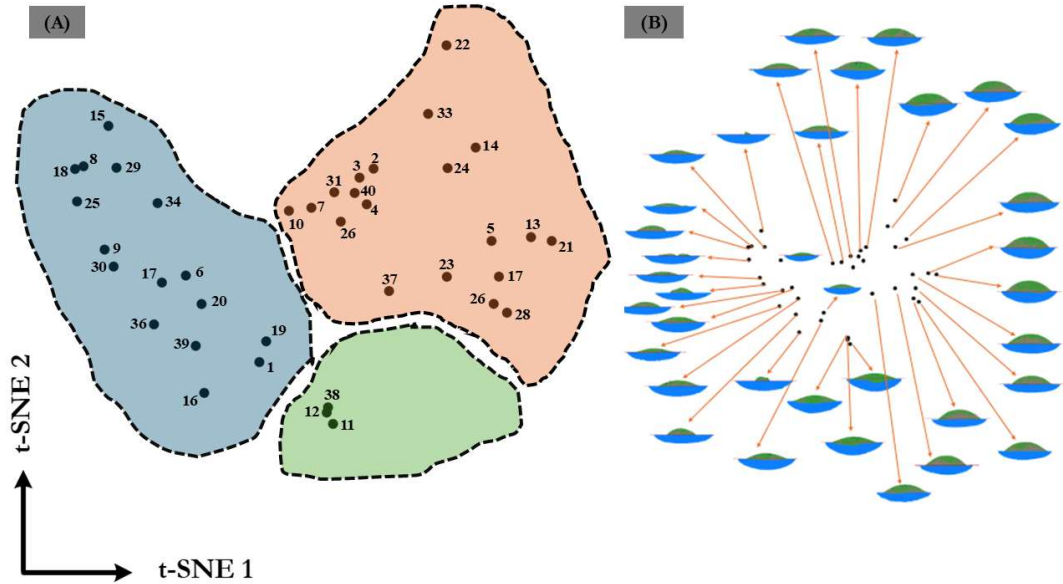

Figure 9: Pretrained vision space of FLAVA model represented with metallographic dataset. (a) identified clusters, and (b) dataset distribution. The numbers in (a) correspond to sample numbers in the dataset.

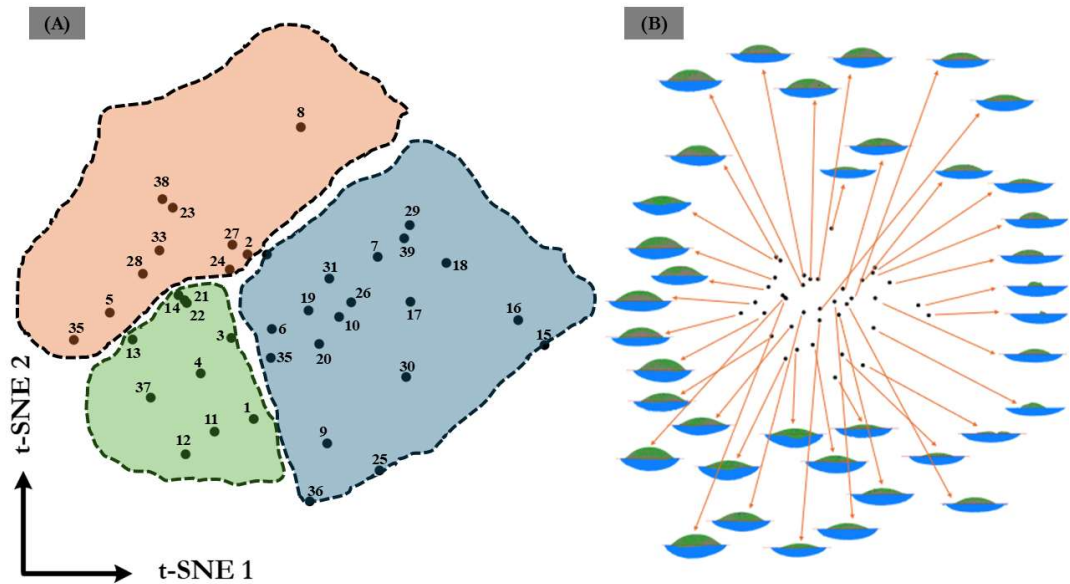

Figure 10: Pretrained vision space of CLIP model represented with metallographic dataset, (a) identified clusters, and (b) dataset distribution. The numbers in (a) correspond to sample numbers in the dataset.

In contrast, the blue region corresponds to lower bead height or complete failure. Excessive dilution cases form a distinct small cluster highlighted as the green region. The clustering behaviour is promising for several reasons. First, both models relatively capture the variations in the microstructural images. Second, both models have differences, which are represented by changes in the pre-training processes and associated datasets. These differences could provide balancing strengths. Finally, owing to its strong potential, the zero-shot classification can be investigated prior to the fine-tuning considerations.

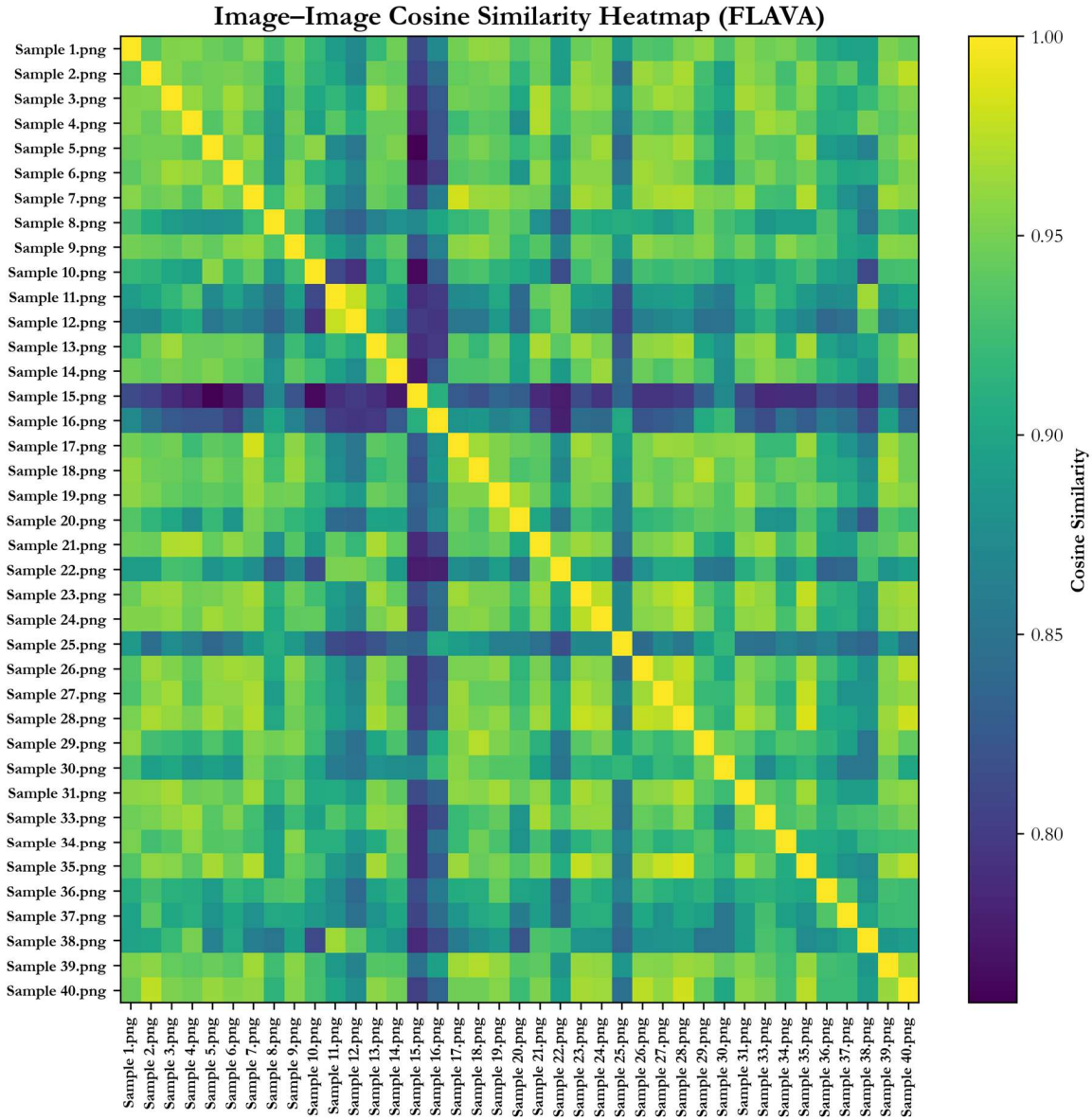

Figure 11: Cosine similarity of FLAVA based vision embeddings across the dataset

A closer examination of individual samples further illustrates how the models reflect key microstructural differences, as shown in respective CLIP (Figure 12) and FLAVA (Figure 11) image-image similarity heatmaps. Samples 15 and 16, which have minimal bead reinforcement areas, display low similarity with most other samples, indicating that both models can distinguish these outliers based on limited reinforcement geometry. Moreover, samples (e.g., Sample 25) that exhibit complete detachment failure also show weak similarity across the dataset. By contrast, samples (e.g., Sample 11 and Sample 12) that are characterized by significant dilution lead to high mutual similarity. These observations suggest that the models capture dilution effects as a distinct visual pattern. Samples with regular reinforcement areas and improved carbide reinforcement lead to clusters of high similarity (e.g., Sample 5, Sample 23, Sample 26, Sample 27, and Sample 40). These patterns indicate that the embeddings are not only sensitive to overall structural differences, but also reflect finer distinctions, such as dilution extent and carbide dispersion.

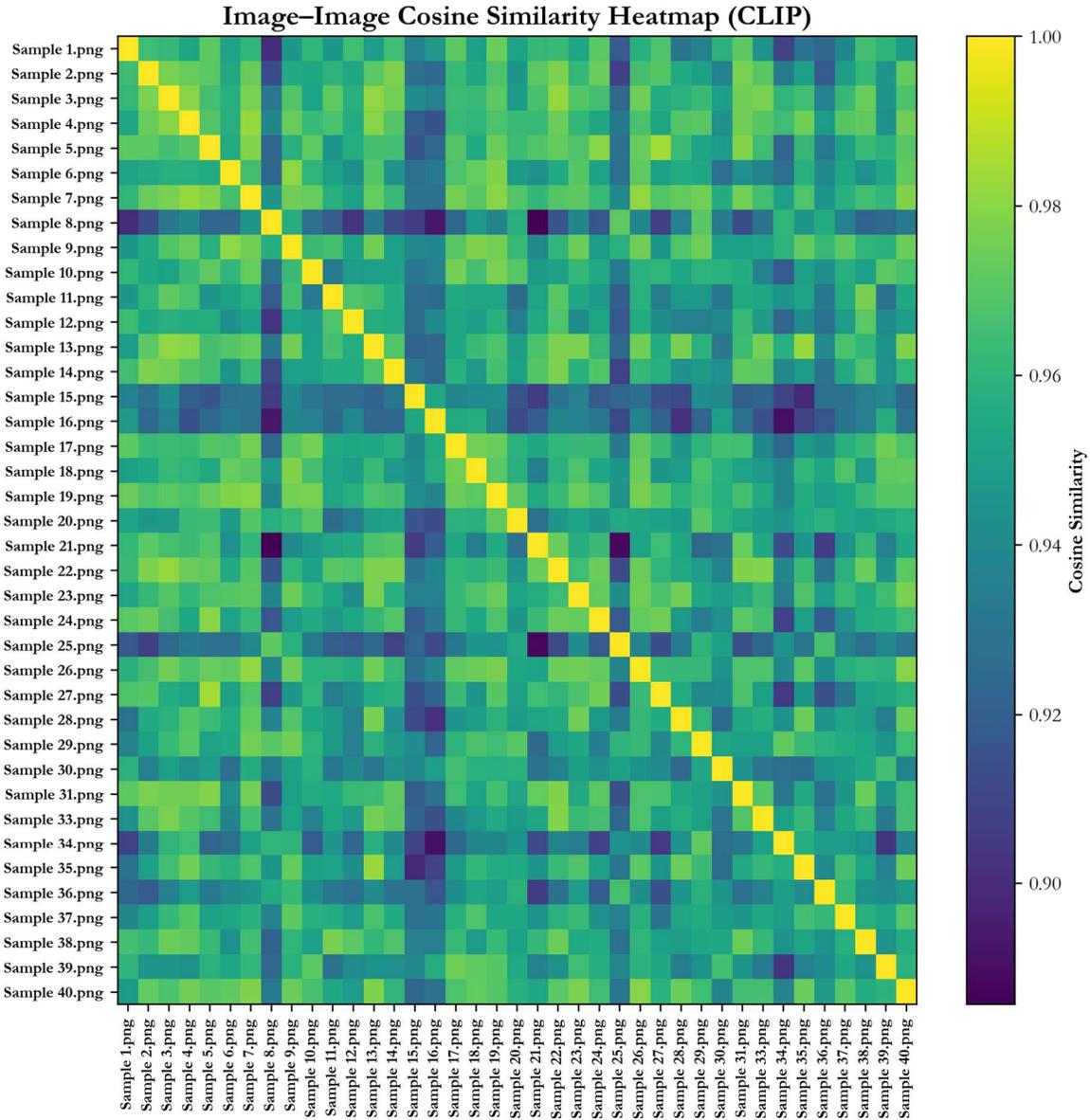

Figure 12: Cosine similarity of CLIP based vision embeddings across the dataset

The difference between FLAVA and CLIP heatmaps provides insight into the behavior of each model when encoding visual similarity. On average, FLAVA produces lower similarity scores compared to CLIP, as reflected in a mean difference of -0.038 (standard deviation: 0.038). However, this reduction is not uniform. FLAVA assigns higher scores to microstructures that are more visually similar and lower scores to dissimilar microstructures. The similarity behavior reflects higher sensitivity to visual differences. For instance, Sample 7 and Sample 8 with similar process parameters have positive differences due to FLAVA's capacity to effectively capture local continuity. In contrast, pairs such as Sample 14 and Sample 25 show large negative differences (e.g., as low as -0.167), implying that FLAVA suppresses weak or spurious similarity that CLIP may overestimate. These findings suggest that FLAVA produces more discriminative representations by enhancing contrast between similar and dissimilar microstructures, which may be beneficial for downstream tasks requiring fine-grained visual reasoning.

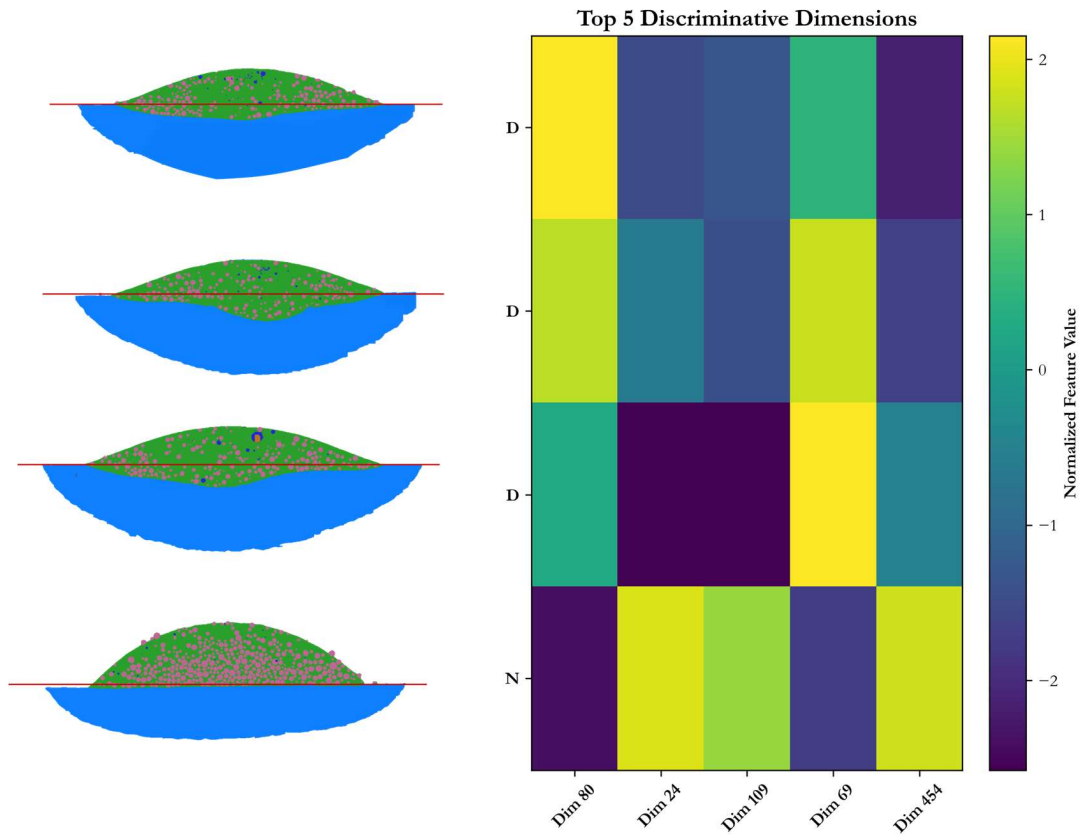

Figure 13: Variation in CLIP image embedding with and without dilution. The heatmap and line plot highlight top discriminative dimensions between defective (D) and Normal (N) examples using standardized feature values

To examine whether the learned visual embeddings encode interpretable representations aligned with domain knowledge, we analyzed how individual feature dimensions vary with changes in key microstructural characteristics. Specifically, two sets of samples were used: one characterized by dilution (green reinforcement extending below the ideal red fusion line) and the other by differences in carbide distribution within the green bead reinforcement area. The first analysis, performed using CLIP's vision encoder, highlights the top five discriminative dimensions that separate diluted and dilution-free samples.

As shown in the heatmap of Figure 13, the feature values of dimensions 80 and 69 clearly differentiate the dilution-free sample (bottom row, labeled "N") from the diluted samples above it.

In the second analysis, performed using FLAVA's vision encoder, we isolated samples with varying carbide distribution patterns and grouped them into uniformly distributed (UD) and non-uniform (ND) categories. The heatmap on the top-right of Figure 14 reveals a different set of discriminative dimensions, including dimension 4 and dimension 346, which exhibit lower feature values for uniformly distributed samples. These analyses validate that both models encode specific knowledge components relevant to expert assessments. While CLIP more clearly separates samples based on dilution, FLAVA demonstrates sensitivity to subtle distribution patterns in the reinforcement zone. These findings suggest that interpretable embedding dimensions may be linked to key microstructural indicators, enabling downstream reasoning tasks without explicit labels.

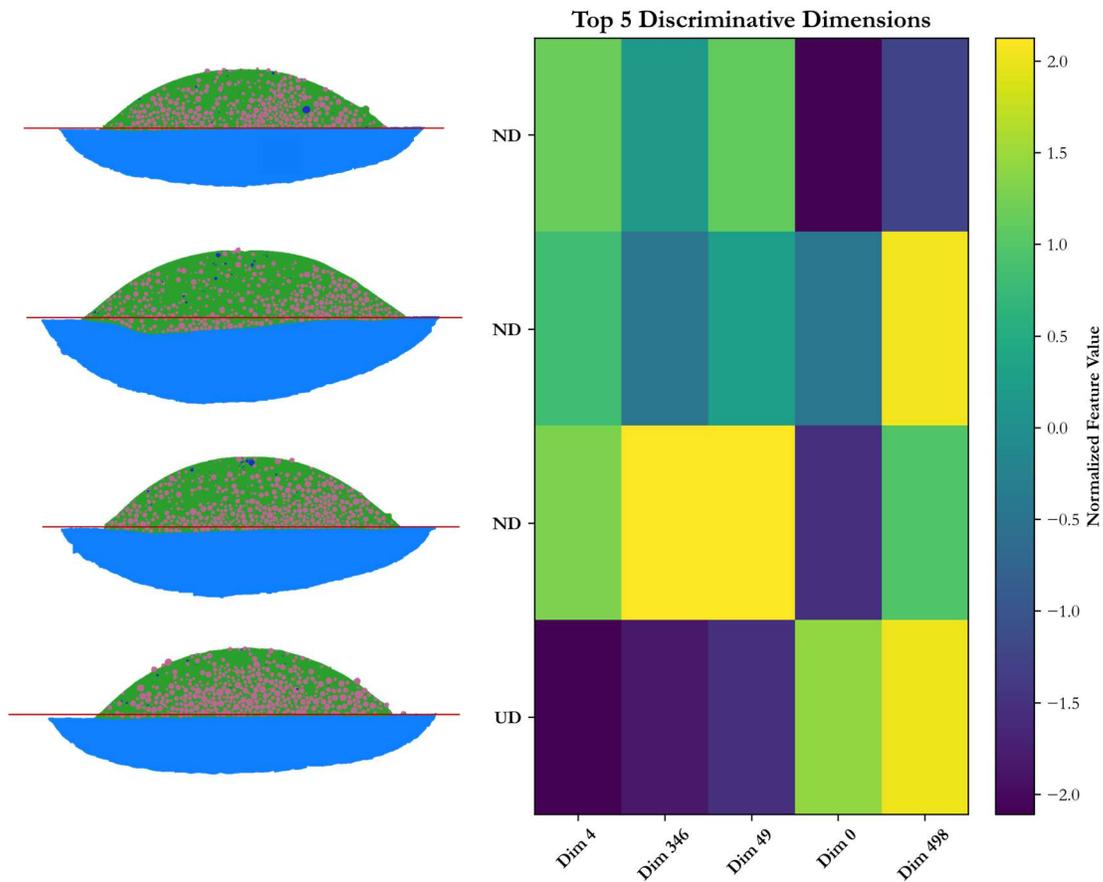

Figure 14: Variation in FLAVA image embeddings across varying carbide distributions. The heatmap and line plot highlight top discriminative dimensions between Non-uniform (ND) and near Uniform (UD) examples using standardized feature values

The accompanying line plots in Figure 15 show how highlighted dimensions exhibit consistent shifts across the dilution and distribution categories, confirming their alignment with target structural variation.

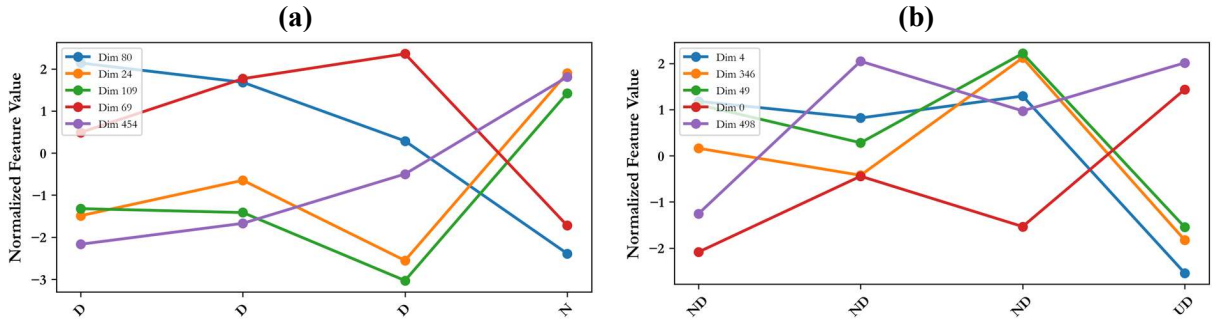

Figure 15: Feature value variations for normal vs diluted samples (a) and non-uniform vs uniform distribution samples

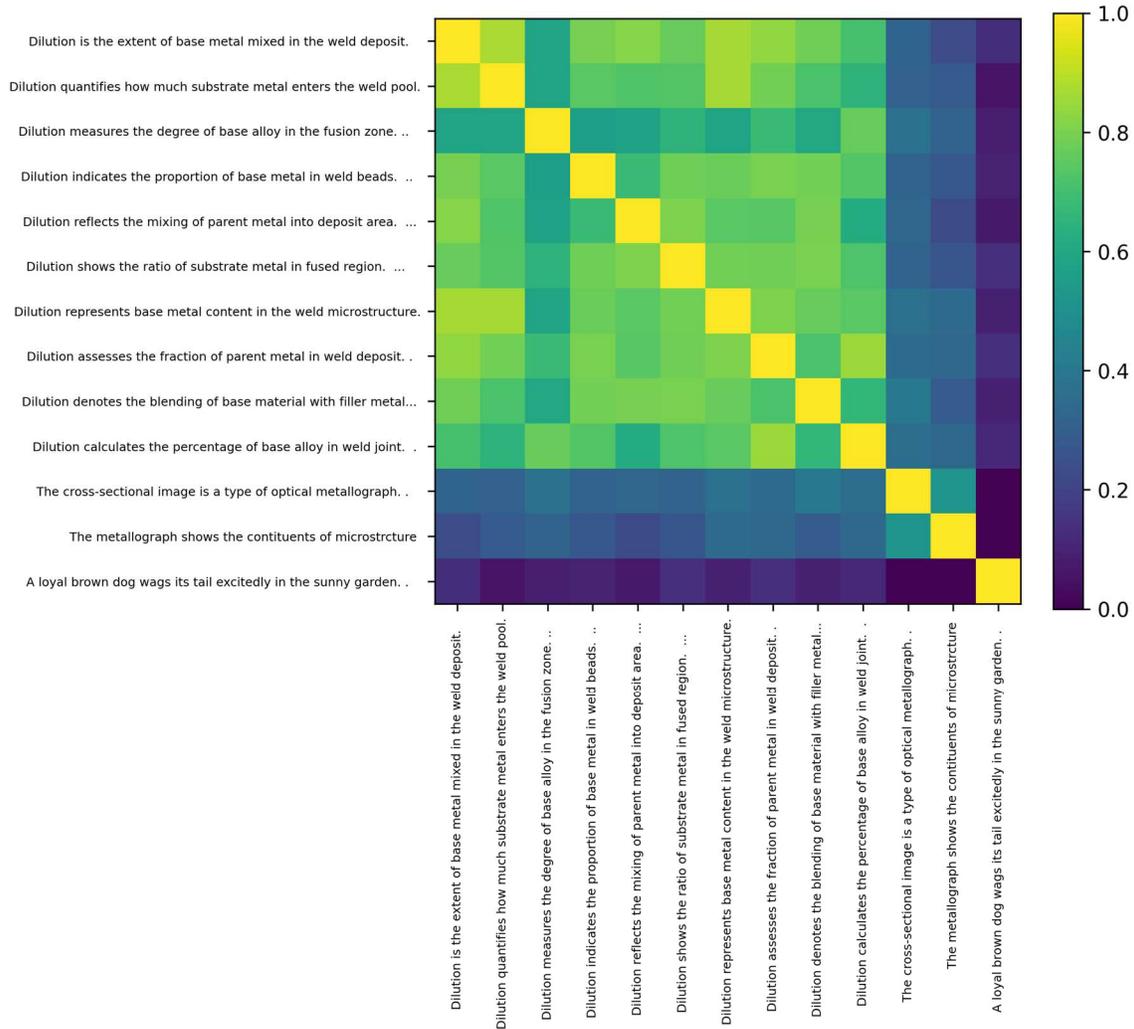

Figure 16: Text similarity plots using FLAVA and example dilution descriptions

Textual descriptions of dilution assessment were used to evaluate the usefulness of text embeddings generated by the two models. Each description offered a slightly different phrasing of the same concept, ranging from precise technical definitions to more generic formulations. Additionally, a baseline sentence representing the generic computer vision domain ("A loyal brown dog...") was included to evaluate out-of-domain similarity. The resulting similarity matrices, shown in Figure 16 (FLAVA) and Figure 17 (CLIP), reveal contrasting patterns across the two models. CLIP exhibits higher intra-class similarity among the dilution-related sentences, suggesting that its text encoder maintains a stronger alignment across paraphrased domain content. Even as descriptions shift from specific (e.g., "extent of base metal mixed in the weld deposit") to more generic language (e.g., "base alloy in weld joint"), the similarity remains relatively high. By contrast, FLAVA's text encoder shows a steeper drop-off in similarity as the descriptions become more abstract or reworded, indicating increased sensitivity to surface-level textual variation.

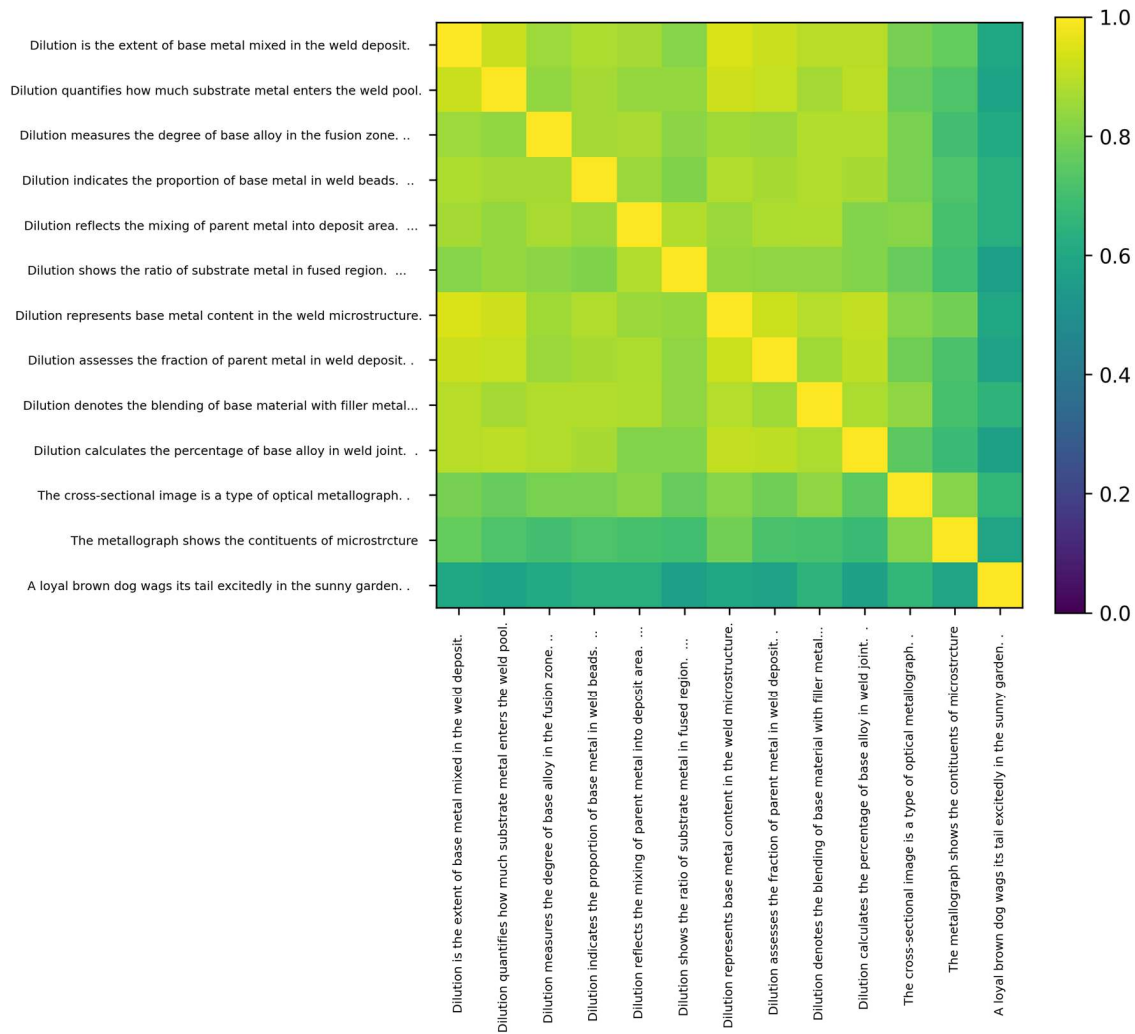

Figure 17: Text similarity plots using CLIP and example dilution descriptions

The observed behavior of both models on textual expert assessments underscores an important difference. CLIP appears better suited for capturing high-level semantic equivalence in domain-relevant language. At the same time, FLAVA places greater weight on token-level differences and is more suitable for joint multi-modal reasoning. The trend was further visualized in a CLIP-FLAVA difference heatmap, where most entries were positive. These results confirmed that CLIP assigns consistently higher similarity scores across related expert assessments. This contrast is prominent among sentences that vary in phrasing but convey the same technical concept, which highlights CLIP's relative robustness to text variation. Notably, both models appropriately assign low similarity to the unrelated baseline sentence. Therefore, these models can differentiate domain-specific knowledge from generic knowledge. Together, these findings suggest that CLIP's text encoder may provide a more stable embedding space for capturing domain expertise expressed in varied textual representations. At the same time, FLAVA may benefit from domain-adaptive tuning to achieve comparable alignment.

Before customizing the extracted embeddings, the image-to-text linkages were evaluated using the selected pre-trained models. Figure 18 shows the dataset in a reduced vision space from each model using t-SNE reduced vision encodings of the high-dimensional space. This distribution of reduced features changes as textual features of the assessments are plotted alongside the vision features. This is reflected in the case of FLAVA (Figure 19) and CLIP (Figure 20) models. While all assessments were plotted, the discussion is limited to textual representation of distribution assessments. The plots highlight a form of knowledge pull (for similar microstructures) and knowledge push (for different microstructures) where the 2D arrangement of image features gets re-distributed according to the image-to-text similarities. In the case of FLAVA, we can see that the text embedding is placed closer to Samples 27 and 35, which contain a uniform distribution of carbide particles. Similarly, in the case of FLAVA, the two samples placed closer to the text assessment describing acceptable carbide distribution in the reinforcement area are Samples 28 and 35. Notably, sample 35 is closer to text embeddings from both models, which may be due to its near-ideal distribution of carbide particles.

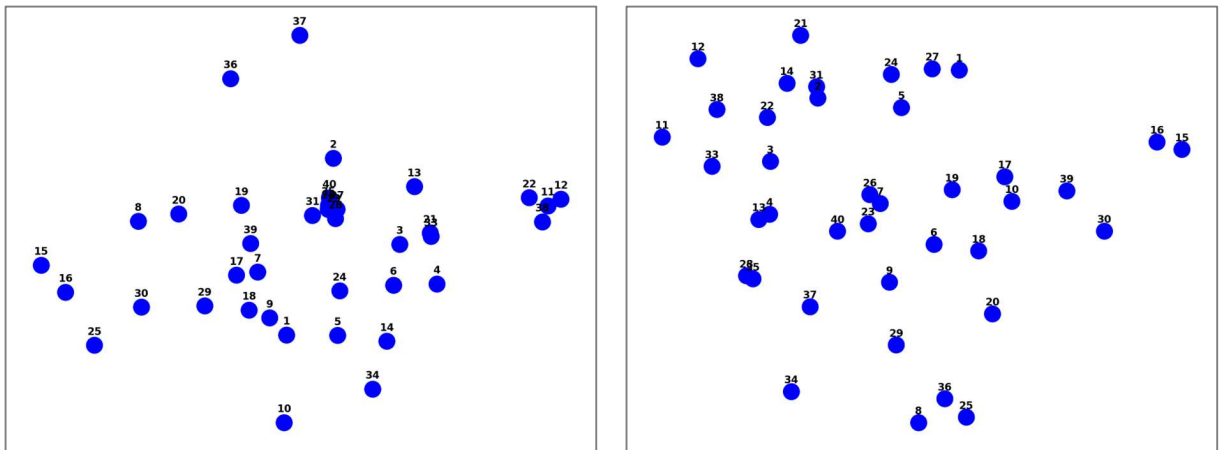

Figure 18: FLAVA (right) and CLIP (left) initial image-only distribution in reduced t-SNE 2D space

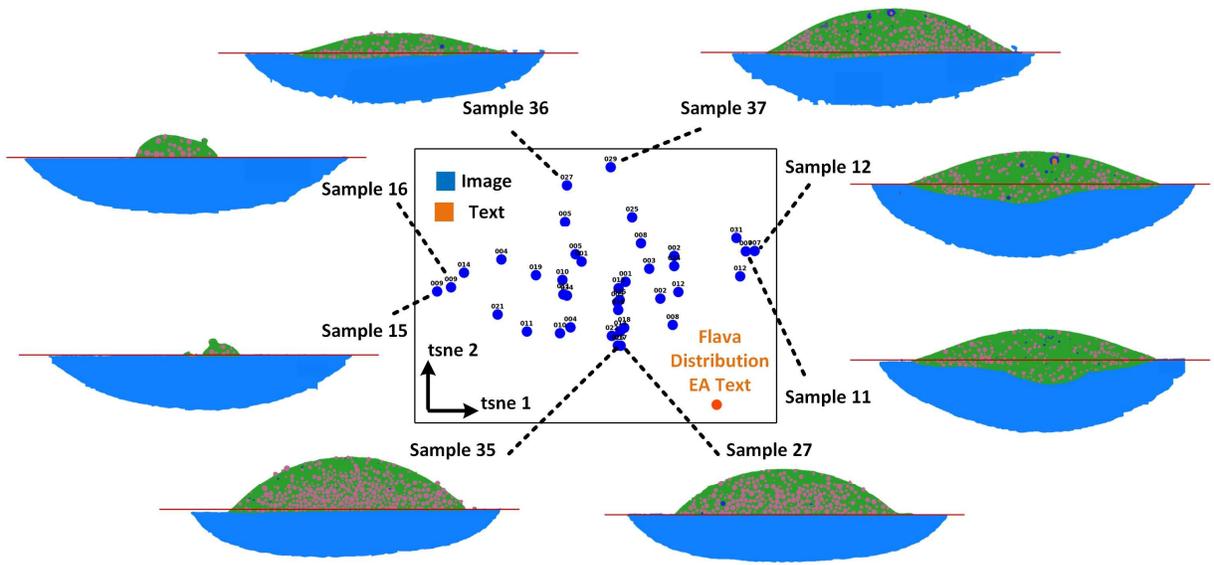

Figure 19: FLAVA shared vision-language distribution in t-SNE reduced 2D space with a positive carbide distribution EA text

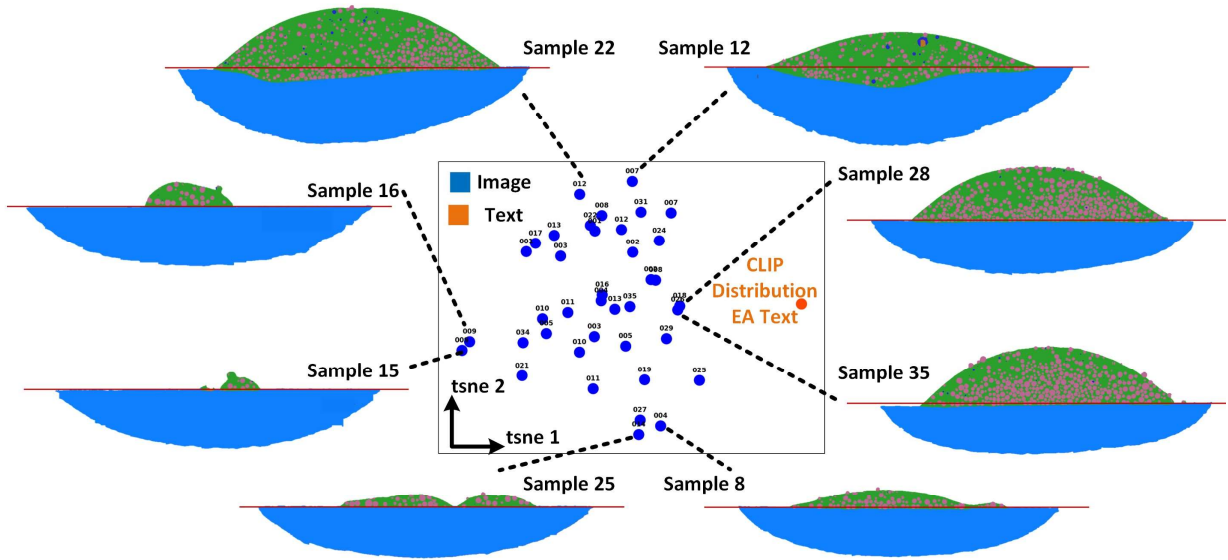

Figure 20: CLIP shared vision-language distribution in t-SNE reduced 2D space with a positive carbide distribution EA text

The image-to-text similarities were quantitatively evaluated using six individual embeddings of expert assessments and combining them successively through averaging. This was accomplished using a combination of cosine similarities and top-k-ranked samples. Across the FLAVA-based expert assessments (Figure 21), Sample 28 consistently emerges as the most similar image in both individual and cumulative EA texts, which highlights its strong alignment with all expert descriptions. Other samples show more dynamic behavior. For instance, Sample 4 ranks among the top five individually, but rises to second place across all five cumulative assessments, indicating that the averaging process strengthens its alignment with the overall expert knowledge. Similarly, images, such as Sample 31 and Sample 40, appear frequently in the top ranks, but exhibit improved stability and agreement in the cumulative view. This shift indicates that averaging the textual representations limits inter-assessment variability and leads to shared semantic representations.

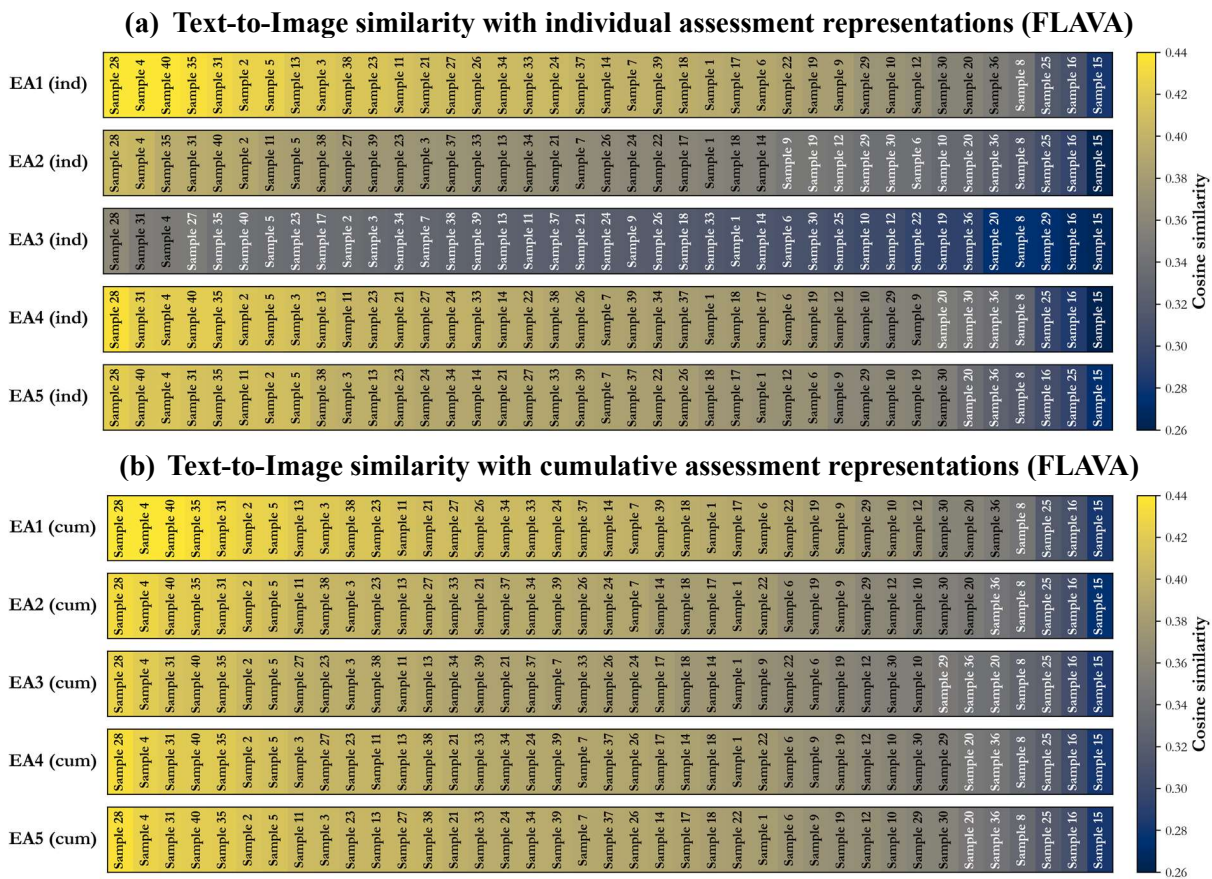

Figure 21: FLAVA text-to-image similarities with individual and cumulative expert assessments. 'ind' references to individual, whereas 'cum' refers to cumulative EA texts

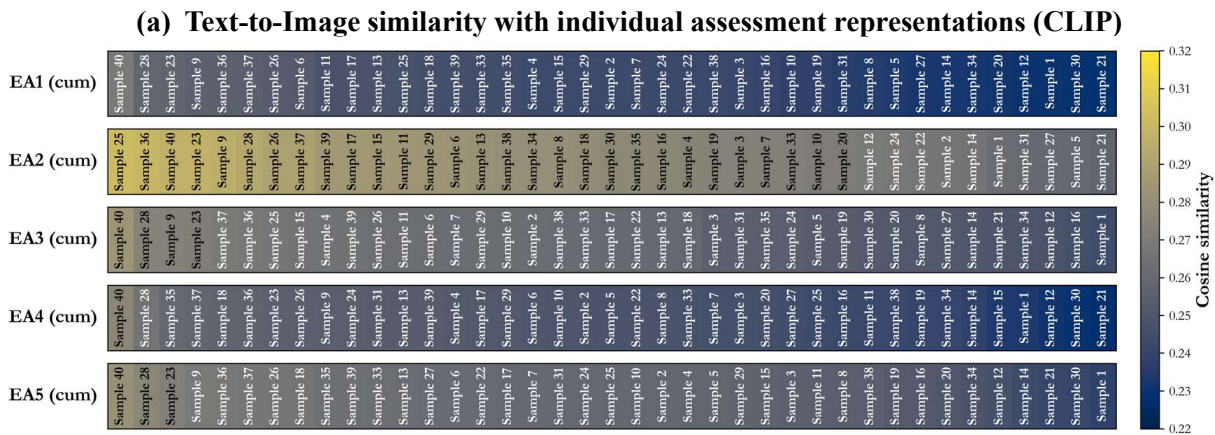

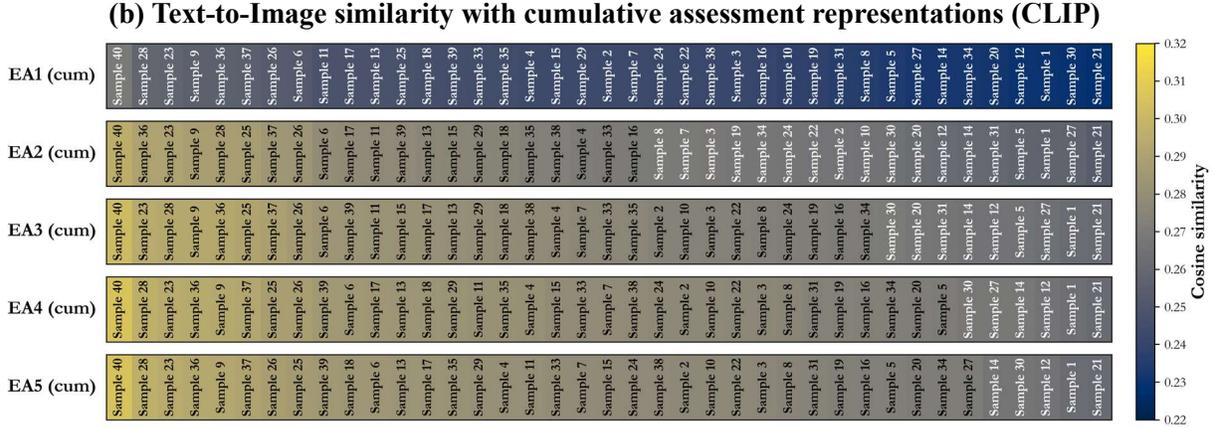

Figure 22: CLIP text-to-image similarities with individual and cumulative expert assessments. 'ind' references to individual, whereas 'cum' refers to cumulative EA texts

In the case of CLIP-based image-to-text similarities (Figure 22), Sample 40 consistently appears as the top-ranked image across all cumulative CLIP assessments, confirming its strong alignment with the averaged textual embeddings. Notably, Sample 25 achieves the highest similarity score in the second individual assessment (EA2 ind, 0.2968), slightly surpassing Sample 40 (0.2952). However, once the generic embedding is averaged in the cumulative version (EA2 cum), Sample 40 regains the lead with a similarity of 0.2961, pushing Sample 25 to sixth place. This pattern recurs in subsequent assessments: Sample 40's similarity score improves from 0.2765 in EA4 (ind) to 0.3059 in EA4 (cum) and from 0.2818 in EA5 (ind) to 0.3053 in EA5 (cum). Meanwhile, Sample 24 moves from third in EA3 (ind, 0.2734) to second in EA3 (cum, 0.2933).

Overall, FLAVA and CLIP display distinct behaviors in terms of both similarity magnitude and ranking consistency. FLAVA embeddings produce noticeably higher cosine similarity scores across all assessments. In the individual setting, Sample 28 dominates every five expert assessments with values ranging from 0.3640 to 0.4457 (mean 0.4146). This behavior continues in the cumulative FLAVA results, where Sample 28 consistently remains on top, and its similarity scores further concentrate within a narrower, elevated band (0.4267–0.4457, mean 0.4348). In contrast, CLIP yields lower similarity values overall. In the individual CLIP assessments, Sample 40 ranks highest in four out of five cases, while Sample 25 briefly leads in the second assessment (0.2968). Once the cumulative averaging is applied, Sample 40 consolidates its dominance with scores increasing and stabilizing between 0.2695 and 0.3059 (mean 0.2957).

Color information on the phases in the microstructural images was added to the textual assessments, and its impact on CLIP-based image-text similarities was evaluated as shown in Figure 23. The information was added directly in the textual prompts to represent each color (e.g., "green matrix", "pink carbides"). This addition led to an increase in the overall cosine similarity scores and a consolidation of top-ranked predictions. For example, in the first individual assessment (EA1), the similarity of sample 40 increased from 0.2695 (without color) to 0.3598 (with color), marking a jump of +0.0903. Similarly, in EA2, Sample 25 initially leads at 0.3024 but is surpassed by Sample 40 once the color is added, reaching a similarity of 0.3601. In the remaining assessments (EA3-EA5), Sample 40 remains dominant, with similarity improvements ranging from +0.0705 to +0.0760. In cumulative evaluations, the gains were consistent: Sample 40's cumulative similarity increases from 0.2695 to 0.3598 in EA1 ($\Delta$ = +0.0903) and from 0.3053 to 0.3323 in EA5 (Δ = +0.0270). These results highlight that integrating color-sensitive textual cues enhances image alignment with expert assessments, particularly for top-ranked predictions.

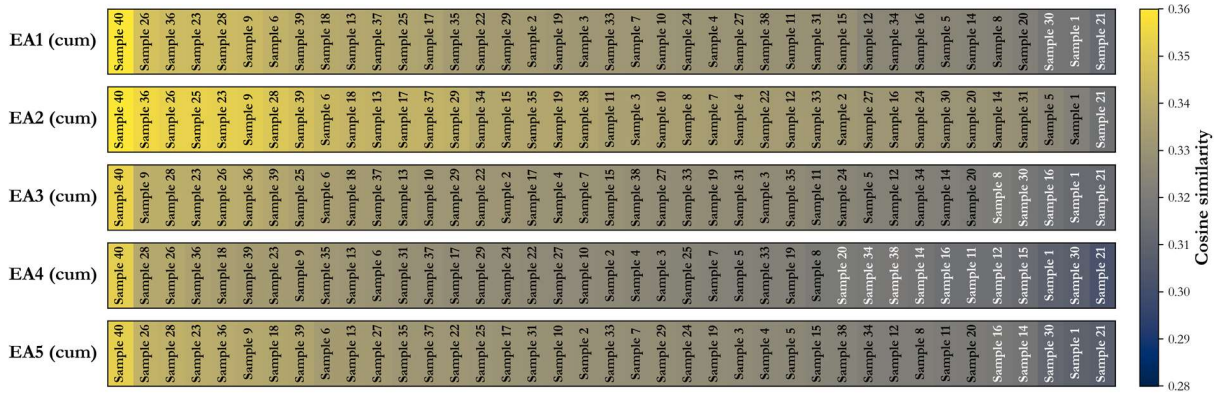

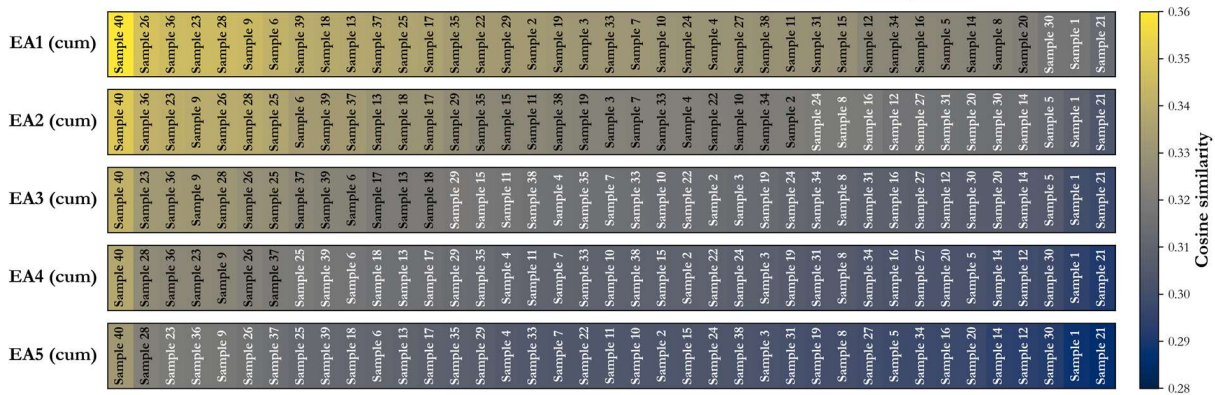

Figure 23: Color-aware expert assessments for text-to-image similarities in CLIP. 'ind' references to individual, whereas 'cum' refers to cumulative EA texts

When comparing individual EAs to image samples using cosine similarity, both FLAVA and CLIP demonstrate distinct behaviors in their retrieval performance. The results are presented using top ranked accuracies for individual (Table 2) as well as cumulative (Table 3) assessments generated by the two models. FLAVA maintains substantial early precision, achieving highest top-5 accuracies of 60% in EA1 (e.g., Dilution), 40% in EA3 (e.g., Porosity), and 80% in EA5 (e.g., Distribution). However, its top-10 and top-15 accuracies decline to 30-40%, suggesting that while FLAVA reliably retrieves a few highly relevant samples, it may be less effective at ranking broader sets of moderately relevant examples. CLIP, in contrast, yields slightly more balanced performance across the top-10 and top-15 ranges, especially in EA3 and EA5, where it reaches 50-60%, indicating better mid-rank retrieval despite more variability in its top-5 predictions. For instance, adding color-aware textual cues in "CLIP (color)" improves overall performance, especially in EA1 and EA5, where top-10 and top-15 accuracies reach 60%. These improvements highlight that incorporating color-sensitive information allows VLMs to align expert textual criteria with visual image features more effectively.

Table 2: Individual FLAVA and CLIP EAs against all samples

| EAs | FLAVA | | | CLIP | | | CLIP (color) | | |
|---|---|---|---|---|---|---|---|---|---|
| | Top-5 Acc | Top-10 Acc | Top-15 Acc | Top-5 Acc | Top-10 Acc | Top-15 Acc | Top-5 Acc | Top-10 Acc | Top-15 Acc |
| #1 | 60% | 30% | 26.67% | **60%** | 40% | 46.67% | 40% | **50%** | **53%** |
| #3 | **40%** | 30% | 33.33% | 20% | 50% | **46.67%** | 20% | **50%** | 40% |
| #5 | 80% | 40% | 40% | 80% | 60% | **60.00%** | 80% | 60% | 53% |

In the cumulative approach, textual descriptions from multiple EAs are progressively averaged to form a more comprehensive embedding for similarity comparison. FLAVA maintains stable performance across the cumulative steps, consistently achieving a top-5 accuracy of 60%, while top-10 and top-15 accuracies fluctuate between 30% and 40%. This suggests that while FLAVA reliably identifies the most confident matches early on, its performance remains inadequate with cumulative textual enrichment-implying that its vision-text alignment is primarily tuned to localized expert descriptors rather than benefiting from semantic aggregation. In contrast, CLIP exhibits noticeable gains in broader retrieval as more assessments are combined. Its top-10 and top-15 accuracies rise to 60% in the final cumulative step (#1-5), indicating enhanced alignment with combined descriptions. However, the most consistent and improved performance is observed in CLIP (color), where top-5 accuracies remain at 60% across all steps, and top-10 and top-15 scores reach 60% by the end. This reflects that CLIP leverages the semantic breadth of cumulative prompts and benefits significantly from color-sensitive alignment. Overall, the results affirm that combining cumulative and color-aware strategies improves retrieval robustness in visually dense tasks, like microstructural interpretation.

Table 3: Cumulative FLAVA and CLIP EAs against all samples

| EAs | FLAVA | | | CLIP | | | CLIP (color) | | |
|---|---|---|---|---|---|---|---|---|---|
| | Top-5 Acc | Top-10 Acc | Top-15 Acc | Top-5 Acc | Top-10 Acc | Top-15 Acc | Top-5 Acc | Top-10 Acc | Top-15 Acc |
| #1 | 60% | 30% | 33.33% | 60% | 50% | 53.33% | **60%** | **60%** | **60%** |
| #1, 2, 3 | 60% | 40% | 33.33% | 60% | 50% | 46.67% | **60%** | 50% | **53.33%** |
| #1,2,3,4,5 | 60% | 30% | 33.33% | 60% | 60% | 60% | **60%** | **60%** | **60%** |

The effectiveness of text customization is limited in single modality without utilizing the image customization for microstructure qualification. In order to fully leverage the freedom to customize the embeddings, we need to also customize the vision space. Jointly, it can help effectively qualify the incoming images of microstructures against multiple criteria by providing reference in both vision and language spaces. As it has been discussed that vision to vision similarity is valid with the pretrained embeddings, their customization is also logical as different examples can cover different aspects of a similar assessment. For instance, example microstructures for distribution and dilution are presented in Figure 24 and Figure 25 respectively. We can see how different variants of the same category upon union can improve the linkage between the two modalities. Upon close examination of the four distribution masks presented, different samples provide varying spatial distribution of the carbide particles. For instance, Sample 5 has an excellent distribution of the reinforcement particles along the fusion line, but the top region of the reinforcement area is missing carbide particles. Sample 27, on the other hand, provides better coverage of the particles in the upper region of the reinforcement area. Similarly, the dilution in the four samples presented in Figure 25 varies in its extent and spatial locality. Therefore, combined vision embeddings of samples with different dilution could provide a better understanding of the overall dilution profile across the sample and act a stronger reference to guide the qualification of future samples.

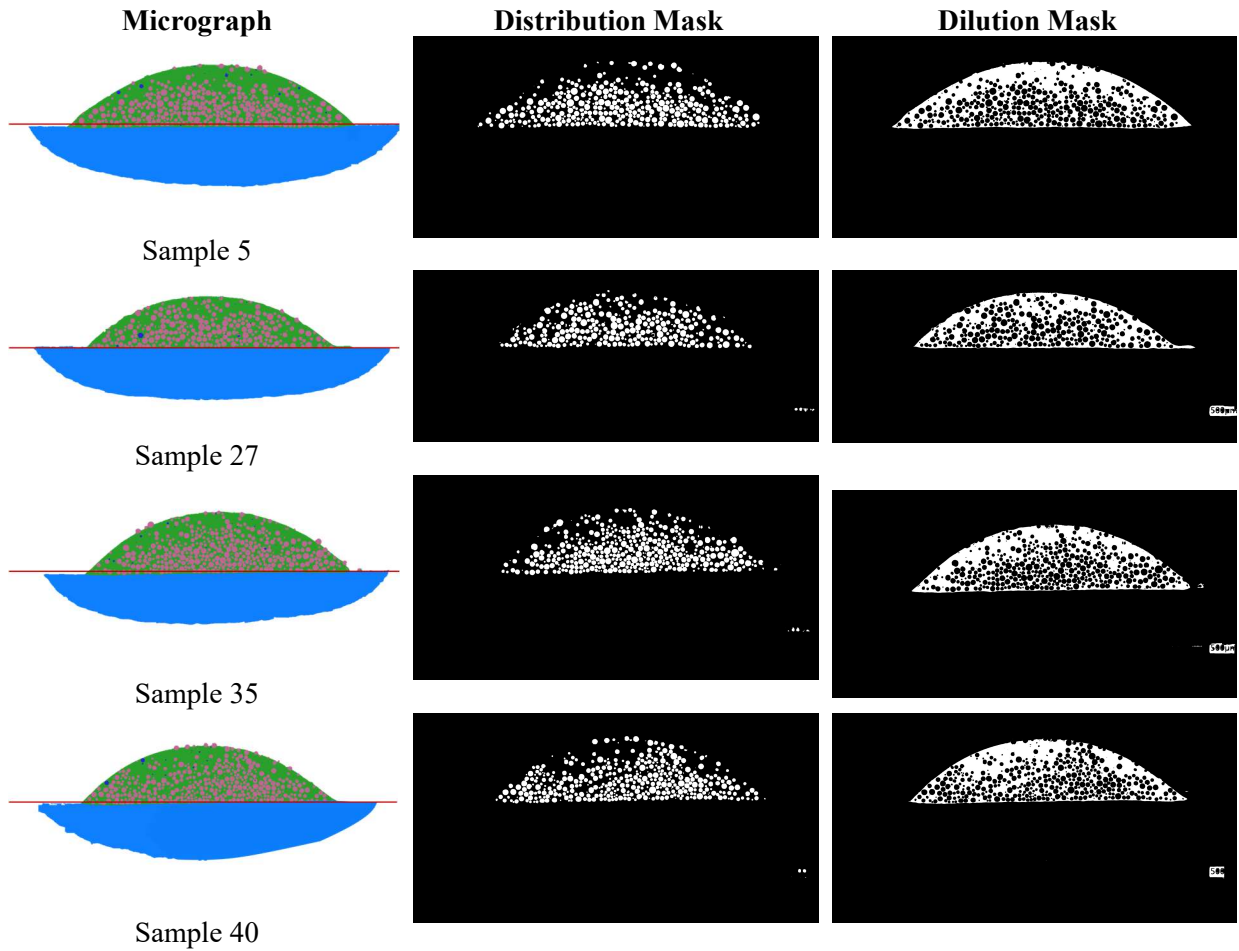

Figure 24: Normal examples of carbide distribution and base dilution across the dataset alongside semantically segmented masks

Following the approach discussed in Section 4, a customized and hybrid method for linking the two modalities was implemented to qualify the incoming test data. We employed a hybrid vision-language similarity framework that integrates features from the FLAVA and CLIP models. First, representative positive and negative image samples were selected and passed through the pre-trained FLAVA model to obtain normalized visual embeddings. Similarly, descriptive positive and negative textual prompts, reflecting ideal and non-ideal reinforcement characteristics, were processed using the CLIP model to generate corresponding text embeddings. The same approach was repeated to get the visual embeddings from the CLIP model, leading to same-size feature vectors suitable for similarity comparison. For each modality, average positive and negative embeddings were computed, and cosine similarity was used to score all images in the dataset with respect to both positive and negative references. These similarity scores were standardized and fused by summing the z-normalized deltas from FLAVA and CLIP, yielding a single combined similarity score for each image. The resulting delta was split at zero to assign binary class labels. Human-labeled annotations for the microstructures were used to evaluate classification performance via confusion matrices and classification metrics. This approach leverages visual features and textual semantics in a multi-modal fashion to enable robust classification of bead reinforcement quality.

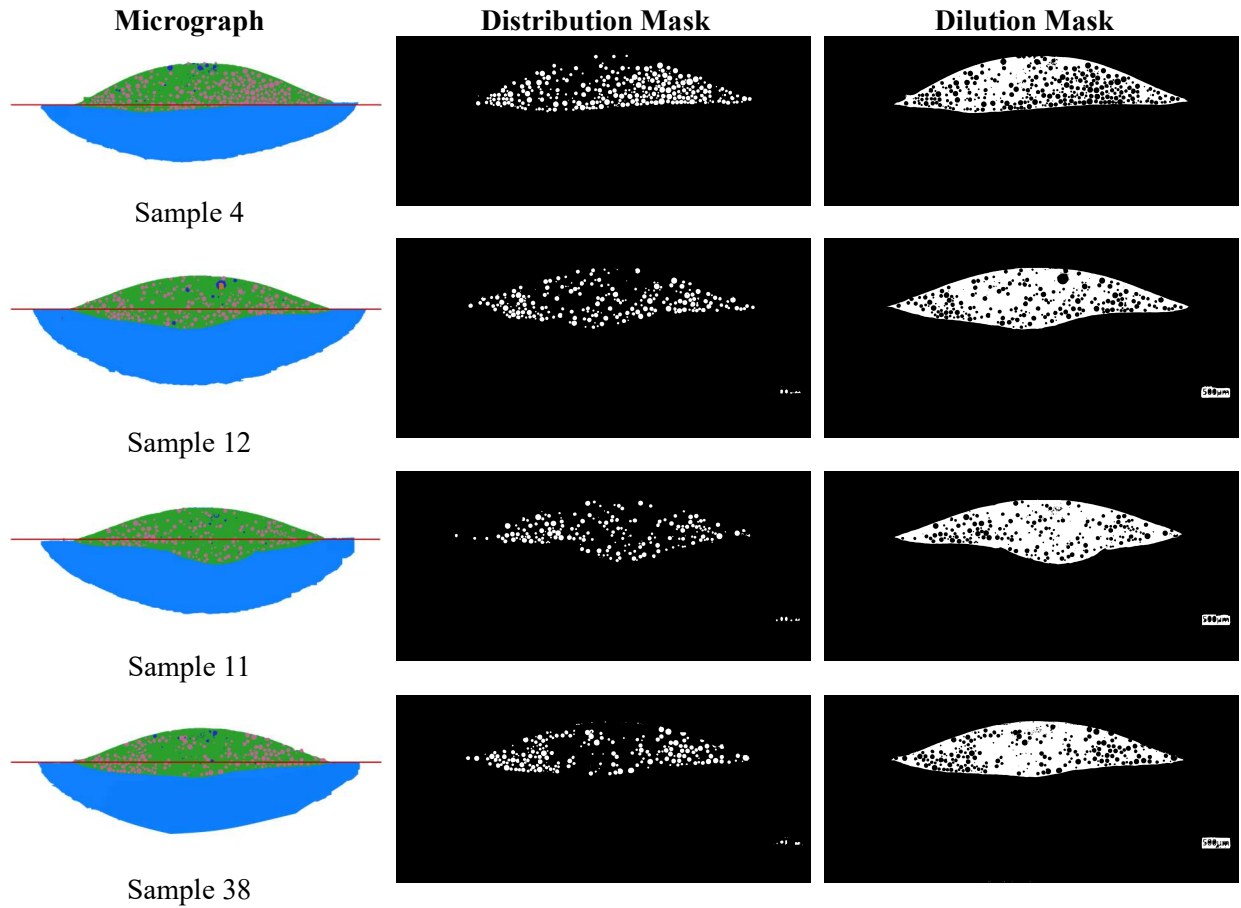

Figure 25: Defective examples of carbide and base dilution across the dataset alongside semantically segmented masks

Figures 26, 27, and 28 present the classification results for distribution, dilution, and reinforcement based expert assessments, respectively. Classification of carbide distribution based on customized and hybrid representations lead to four false positives (samples 17, 18, 25, and 29) and three false negatives (samples 4, 21, and 37). Upon investigation, false positive samples exhibited non-parabolic or a very thin reinforcement area while maintaining a high relative proportion of the carbide particles, as shown in the first part of Figure 29. On the other hand, false negatives tend to come from samples with parabolic reinforcement areas with uniform distribution only in the limited regions of the cross-sections, as shown in the second part of Figure 29.

Referring to the hybrid classification table, the impact of z-score normalization becomes particularly evident near the decision boundary (e.g., zero). Sample 2, for instance, has a negative raw CLIP delta ($-0.0038$), which would ordinarily weaken its classification confidence. However, after normalization, its CLIP z-score becomes positive ($0.1276$), complementing the strong FLAVA z-score ($0.9595$) and yielding a high combined score of $1.0870$. This resulted in a confident positive prediction. By contrast, Sample 3

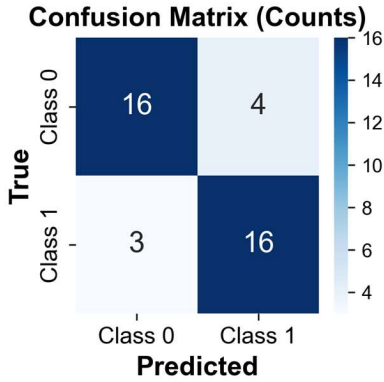

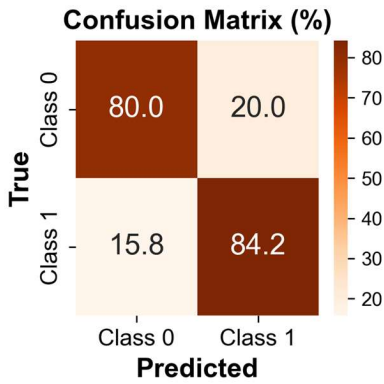

| Image | delta_flava | delta_clip | delta_flava_z | delta_clip_z | delta_combined |
|---|---|---|---|---|---|
| Sample 35.png | 0.1848 | -0.0026 | 1.7307 | 0.6174 | 2.3481 |
| Sample 40.png | 0.1321 | -0.0013 | 1.0835 | 1.2114 | 2.2949 |
| Sample 31.png | 0.1166 | -0.0011 | 0.8934 | 1.2932 | 2.1866 |
| Sample 5.png | 0.1236 | -0.0019 | 0.9795 | 0.9265 | 1.9060 |
| Sample 27.png | 0.1253 | -0.0022 | 1.0005 | 0.8104 | 1.8109 |
| Sample 23.png | 0.1217 | -0.0021 | 0.9564 | 0.8522 | 1.8086 |
| Sample 21.png | 0.0807 | -0.0011 | 0.4534 | 1.2810 | 1.7345 |
| Sample 33.png | 0.0973 | -0.0019 | 0.6568 | 0.9214 | 1.5782 |
| Sample 24.png | 0.1220 | -0.0030 | 0.9598 | 0.4663 | 1.4261 |
| Sample 10.png | 0.0853 | -0.0020 | 0.5091 | 0.9075 | 1.4166 |
| Sample 7.png | 0.1103 | -0.0027 | 0.8159 | 0.5753 | 1.3913 |
| Sample 26.png | 0.1663 | -0.0045 | 1.5039 | -0.1850 | 1.3189 |
| Sample 37.png | 0.0813 | -0.0025 | 0.4600 | 0.6920 | 1.1519 |
| Sample 2.png | 0.1220 | -0.0038 | 0.9595 | 0.1276 | 1.0870 |
| Sample 28.png | 0.1100 | -0.0037 | 0.8127 | 0.1572 | 0.9700 |
| Sample 20.png | 0.0993 | -0.0035 | 0.6817 | 0.2711 | 0.9528 |
| Sample 39.png | 0.0416 | -0.0020 | -0.0265 | 0.8961 | 0.8696 |
| Sample 8.png | -0.0737 | 0.0013 | -1.4416 | 2.2950 | 0.8533 |
| Sample 4.png | 0.0773 | -0.0032 | 0.4119 | 0.3710 | 0.7829 |
| Sample 13.png | 0.1015 | -0.0047 | 0.7085 | -0.2540 | 0.4544 |
| Sample 14.png | 0.0960 | -0.0057 | 0.6402 | -0.6731 | -0.0328 |
| Sample 3.png | 0.0772 | -0.0059 | 0.4099 | -0.7706 | -0.3607 |
| Sample 30.png | -0.0260 | -0.0030 | -0.8559 | 0.4751 | -0.3808 |
| Sample 22.png | 0.0545 | -0.0054 | 0.1317 | -0.5429 | -0.4111 |
| Sample 19.png | 0.0516 | -0.0057 | 0.0963 | -0.7012 | -0.6049 |
| Sample 32.png | -0.0078 | -0.0041 | -0.6333 | -0.0096 | -0.6429 |
| Sample 36.png | -0.0190 | -0.0038 | -0.7703 | 0.1112 | -0.6591 |
| Sample 6.png | 0.0227 | -0.0054 | -0.2582 | -0.5648 | -0.8230 |
| Sample 29.png | -0.0498 | -0.0034 | -1.1481 | 0.3104 | -0.8376 |
| Sample 34.png | -0.0248 | -0.0041 | -0.8410 | -0.0085 | -0.8495 |
| Sample 17.png | 0.0145 | -0.0052 | -0.3596 | -0.4990 | -0.8585 |
| Sample 1.png | 0.0323 | -0.0061 | -0.1403 | -0.8695 | -1.0098 |
| Sample 16.png | 0.0271 | -0.0067 | -0.2049 | -1.1151 | -1.3200 |
| Sample 25.png | -0.0633 | -0.0041 | -1.3143 | -0.0183 | -1.3326 |
| Sample 18.png | -0.0626 | -0.0048 | -1.3051 | -0.3226 | -1.6276 |
| Sample 38.png | -0.0207 | -0.0063 | -0.7911 | -0.9702 | -1.7614 |
| Sample 9.png | -0.0467 | -0.0075 | -1.1107 | -1.4692 | -2.5799 |
| Sample 12.png | -0.0279 | -0.0099 | -0.8797 | -2.4926 | -3.3723 |
| Sample 15.png | -0.1837 | -0.0073 | -2.7916 | -1.3789 | -4.1706 |
| Sample 11.png | -0.1178 | -0.0104 | -1.9832 | -2.7243 | -4.7075 |

Figure 26: Distribution classification results alongside raw, standardized, and combined similarity scores

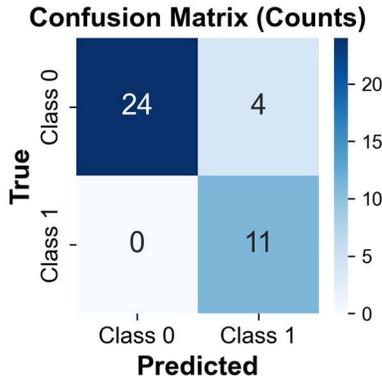
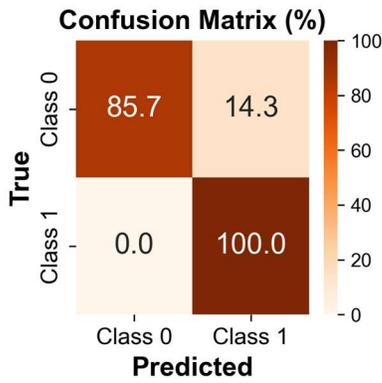

| Image | delta_flava | delta_clip | delta_flava_z | delta_clip_z | delta_combined |
|---|---|---|---|---|---|
| Sample 40.png | 0.1175 | -0.0080 | 0.7308 | 2.1247 | 2.8555 |
| Sample 7.png | 0.1222 | -0.0090 | 0.7953 | 1.6110 | 2.4063 |
| Sample 26.png | 0.1749 | -0.0109 | 1.5168 | 0.6212 | 2.1380 |
| Sample 35.png | 0.1645 | -0.0106 | 1.3740 | 0.7594 | 2.1333 |
| Sample 32.png | 0.0340 | -0.0072 | -0.4111 | 2.5187 | 2.1076 |
| Sample 28.png | 0.0798 | -0.0086 | 0.2153 | 1.7931 | 2.0084 |
| Sample 20.png | 0.1637 | -0.0115 | 1.3626 | 0.2802 | 1.6428 |
| Sample 29.png | 0.1027 | -0.0100 | 0.5282 | 1.0923 | 1.6205 |
| Sample 8.png | 0.0834 | -0.0102 | 0.2636 | 0.9993 | 1.2629 |
| Sample 37.png | 0.0775 | -0.0105 | 0.1829 | 0.8337 | 1.0166 |
| Sample 10.png | 0.1130 | -0.0115 | 0.6699 | 0.3126 | 0.9825 |
| Sample 36.png | 0.0971 | -0.0116 | 0.4516 | 0.2489 | 0.7004 |
| Sample 25.png | 0.0884 | -0.0117 | 0.3329 | 0.1823 | 0.5152 |
| Sample 18.png | 0.0610 | -0.0110 | -0.0420 | 0.5536 | 0.5116 |
| Sample 17.png | 0.1127 | -0.0125 | 0.6651 | -0.1909 | 0.4742 |
| Sample 5.png | 0.0706 | -0.0116 | 0.0885 | 0.2588 | 0.3474 |
| Sample 16.png | 0.1083 | -0.0126 | 0.6056 | -0.2750 | 0.3306 |
| Sample 39.png | 0.1078 | -0.0127 | 0.5978 | -0.2945 | 0.3033 |
| Sample 31.png | 0.1027 | -0.0125 | 0.5284 | -0.2270 | 0.3014 |
| Sample 2.png | 0.0897 | -0.0122 | 0.3510 | -0.0510 | 0.2999 |
| Sample 30.png | 0.1008 | -0.0126 | 0.5020 | -0.2480 | 0.2540 |
| Sample 24.png | 0.0455 | -0.0112 | -0.2544 | 0.4791 | 0.2246 |
| Sample 33.png | -0.0062 | -0.0098 | -0.9614 | 1.1702 | 0.2088 |
| Sample 23.png | 0.1093 | -0.0132 | 0.6181 | -0.5636 | 0.0545 |
| Sample 27.png | 0.0838 | -0.0126 | 0.2704 | -0.2451 | 0.0253 |
| Sample 6.png | 0.1046 | -0.0134 | 0.5544 | -0.6868 | -0.1324 |
| Sample 19.png | 0.0695 | -0.0128 | 0.0743 | -0.3914 | -0.3171 |
| Sample 34.png | 0.0968 | -0.0137 | 0.4472 | -0.8318 | -0.3846 |
| Sample 15.png | 0.0982 | -0.0141 | 0.4674 | -1.0297 | -0.5623 |
| Sample 4.png | 0.0179 | -0.0120 | -0.6316 | 0.0445 | -0.5871 |
| Sample 9.png | 0.0898 | -0.0148 | 0.3522 | -1.4031 | -1.0509 |
| Sample 3.png | 0.0233 | -0.0133 | -0.5577 | -0.6036 | -1.1613 |
| Sample 13.png | 0.0423 | -0.0138 | -0.2981 | -0.8839 | -1.1820 |
| Sample 1.png | 0.0135 | -0.0133 | -0.6925 | -0.6403 | -1.3328 |
| Sample 21.png | 0.0050 | -0.0141 | -0.8081 | -1.0207 | -1.8288 |
| Sample 38.png | -0.0184 | -0.0136 | -1.1288 | -0.8039 | -1.9327 |
| Sample 14.png | -0.0070 | -0.0148 | -0.9727 | -1.3799 | -2.3526 |
| Sample 12.png | -0.0371 | -0.0162 | -1.3843 | -2.1271 | -3.5114 |
| Sample 11.png | -0.1258 | -0.0150 | -2.5985 | -1.5191 | -4.1176 |
| Sample 22.png | -0.2140 | -0.0130 | -3.8050 | -0.4673 | -4.2722 |

Figure 27: Dilution classification results alongside raw, standardized, and combined similarity scores

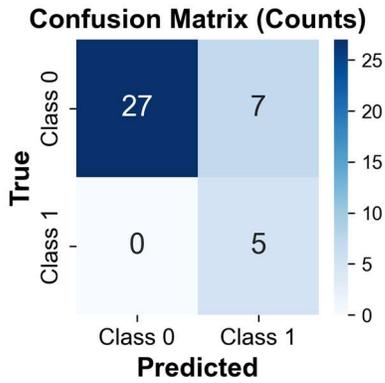

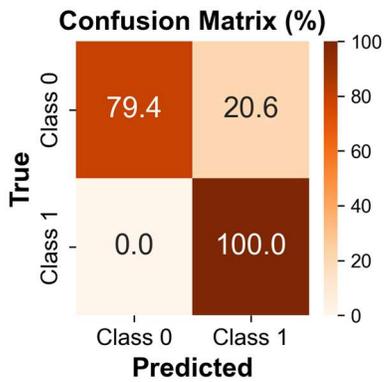

| Image | delta_flava | delta_clip | delta_flava_z | delta_clip_z | delta_combined |
|---|---|---|---|---|---|
| Sample 31.png | 0.0854 | 0.0387 | 0.6028 | 1.9556 | 2.5585 |
| Sample 10.png | 0.0838 | 0.0306 | 0.5729 | 1.1295 | 1.7023 |
| Sample 26.png | 0.1363 | 0.0202 | 1.5235 | 0.0690 | 1.5925 |
| Sample 27.png | 0.0638 | 0.0314 | 0.2104 | 1.2142 | 1.4247 |
| Sample 40.png | 0.0990 | 0.0250 | 0.8489 | 0.5592 | 1.4080 |
| Sample 7.png | 0.0835 | 0.0263 | 0.5689 | 0.6964 | 1.2653 |
| Sample 35.png | 0.0507 | 0.0319 | -0.0264 | 1.2658 | 1.2394 |
| Sample 5.png | 0.0626 | 0.0288 | 0.1892 | 0.9458 | 1.1350 |
| Sample 4.png | 0.0776 | 0.0248 | 0.4608 | 0.5437 | 1.0044 |
| Sample 2.png | 0.0803 | 0.0243 | 0.5096 | 0.4895 | 0.9991 |
| Sample 20.png | 0.0548 | 0.0287 | 0.0488 | 0.9359 | 0.9847 |
| Sample 36.png | 0.1246 | 0.0146 | 1.3121 | -0.5051 | 0.8070 |
| Sample 28.png | 0.0366 | 0.0288 | -0.2813 | 0.9500 | 0.6686 |
| Sample 19.png | 0.0681 | 0.0232 | 0.2891 | 0.3717 | 0.6608 |
| Sample 3.png | 0.0730 | 0.0218 | 0.3776 | 0.2382 | 0.6158 |
| Sample 17.png | 0.0538 | 0.0244 | 0.0310 | 0.4941 | 0.5251 |
| Sample 8.png | 0.0978 | 0.0164 | 0.8265 | -0.3138 | 0.5127 |
| Sample 9.png | 0.0674 | 0.0206 | 0.2761 | 0.1143 | 0.3904 |
| Sample 24.png | 0.0667 | 0.0205 | 0.2630 | 0.1043 | 0.3673 |
| Sample 12.png | 0.0525 | 0.0217 | 0.0059 | 0.2212 | 0.2271 |
| Sample 6.png | 0.0693 | 0.0186 | 0.3117 | -0.0942 | 0.2174 |
| Sample 18.png | 0.0402 | 0.0235 | -0.2168 | 0.4116 | 0.1948 |
| Sample 29.png | 0.0488 | 0.0213 | -0.0602 | 0.1838 | 0.1236 |
| Sample 38.png | 0.0657 | 0.0183 | 0.2459 | -0.1246 | 0.1213 |
| Sample 13.png | 0.0558 | 0.0199 | 0.0663 | 0.0387 | 0.1051 |
| Sample 1.png | 0.0562 | 0.0197 | 0.0746 | 0.0241 | 0.0987 |
| Sample 14.png | 0.0629 | 0.0183 | 0.1959 | -0.1214 | 0.0745 |
| Sample 39.png | 0.0436 | 0.0211 | -0.1545 | 0.1611 | 0.0066 |
| Sample 23.png | 0.0605 | 0.0176 | 0.1509 | -0.1918 | -0.0409 |
| Sample 22.png | 0.0362 | 0.0213 | -0.2885 | 0.1812 | -0.1072 |
| Sample 33.png | 0.0569 | 0.0171 | 0.0871 | -0.2420 | -0.1550 |
| Sample 34.png | 0.0529 | 0.0164 | 0.0136 | -0.3199 | -0.3062 |
| Sample 21.png | 0.0500 | 0.0165 | -0.0386 | -0.3111 | -0.3497 |
| Sample 11.png | 0.0585 | 0.0146 | 0.1158 | -0.5017 | -0.3859 |
| Sample 37.png | 0.0532 | 0.0148 | 0.0202 | -0.4775 | -0.4574 |
| Sample 30.png | 0.0573 | 0.0079 | 0.0934 | -1.1838 | -1.0905 |
| Sample 25.png | 0.0391 | 0.0062 | -0.2366 | -1.3604 | -1.5970 |
| Sample 16.png | -0.0628 | 0.0042 | -2.0812 | -1.5587 | -3.6399 |
| Sample 15.png | -0.0619 | 0.0005 | -2.0647 | -1.9397 | -4.0044 |
| Sample 32.png | -0.2154 | -0.0203 | -4.8439 | -4.0528 | -8.8967 |

Figure 28: Reinforcement area classification results alongside raw, standardized, and combined similarity scores

shows the opposite trend: despite a modest positive FLAVA delta (0.0772), the highly negative CLIP z-score (−0.7706) pulls the combined score below zero (−0.0328), leading to a negative classification. These examples highlight how z-scoring balances each model's influence, allowing weaker signals from one modality to be either amplified or moderated based on cross-modal agreement. Adding to this analysis, z-score normalization from FLAVA flipped the final classification in two cases (e.g., Sample 13, Sample 26), whereas normalization from CLIP was the decisive factor in shifting the outcome in several instances (e.g., Sample 19, Sample 22, Sample 3, Sample 4, Sample 8). There were also cases where both models contributed to correct labelling after normalization (Sample 6). These boundary cases underscore the complementary nature of the hybrid framework, where normalization calibrates each model's raw contribution and hybrid similarities provide a net score to support decision making.

Classification of dilution in the samples based on customized and hybrid representations lead to four false positives (Samples 15, 19, 34, and 9). Upon investigation, the false positive samples exhibited negligible dilution, as shown in Figure 30. Z-score normalization provided similar impact on image-to-text CLIP similarities leading to positive scores for most true positive samples. In this second analysis focused on dilution classification, z-score normalization from FLAVA was responsible for maintaining the classification results in several boundary cases (Sample 16 and Sample 17, Sample 30, Sample 31), where transformed deltas maintained the sign owing to higher contributions from FLAVA. CLIP's standardized contribution was decisive in other cases (Sample 33, Sample 6, and Sample 19), where its higher z-scores offset lower FLAVA contributions, tipping the decision close to the threshold. These shifts illustrate how normalization not only equalizes contribution scales, but also enables subtle yet decisive adjustments near the classification boundary. Notably, all samples with varying extents of dilution were correctly classified, which highlights the effectiveness of providing useful references during the customization process. Moreover, both models supported correct classification in the case of Sample 1, Sample 3, and Sample 13 after standardization.

Lastly, classification of reinforcement area is performed to detect failure due to detachment. This leads to correctly separating all samples with defective reinforcement area. However, seven samples that had the reinforcement area deposited on the substrate got classified as false positives. These misclassified cases represented successful deposition, but significant presence of other defects highlighting the need for references that cover all types of bead reinforcement areas, including defective cases (with dilution, porosities, and non-uniform distribution), in order to correctly classify the presence of reinforcement area.

Figure 31 shows two false positive with defects (excessive dilution as well as porosity) that were missing from the reference embeddings. In this final analysis focused on the hybrid classification of reinforcement areas, z-score normalization from FLAVA flipped the classification for Sample 22, whereas CLIP led to six samples (#11, 23, 30, 33, 34, 37) being assigned negative final score. Both models contributed to correcting the sign of Sample 21 from positive to negative. It is important to note that unlike the previous two analysis, where only two flipped samples represented misclassified cases (Sample 4 in distribution and Sample 19 in dilution), all samples mentioned for CLIP and FLAVA represent the seven false positive cases. Notably, CLIP's contribution in flipping six out of seven samples from positive to negative highlight inadequate semantic reference in the textual domain. Therefore, the reference positive and negative texts need to be augmented by providing knowledge for defective cases where the bead reinforcement area is still present.

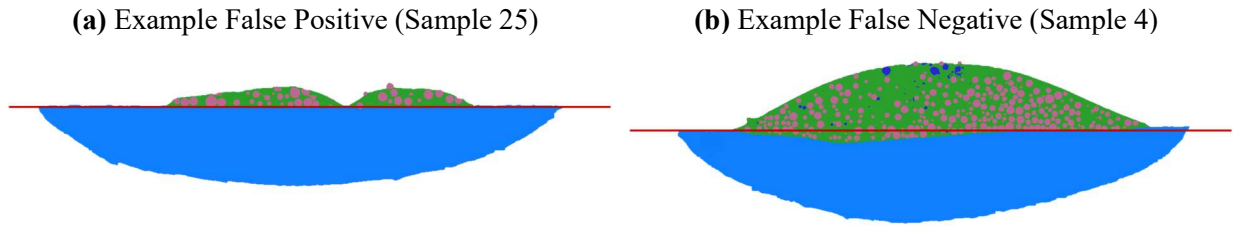

Figure 29: Distribution false positive and false negative examples

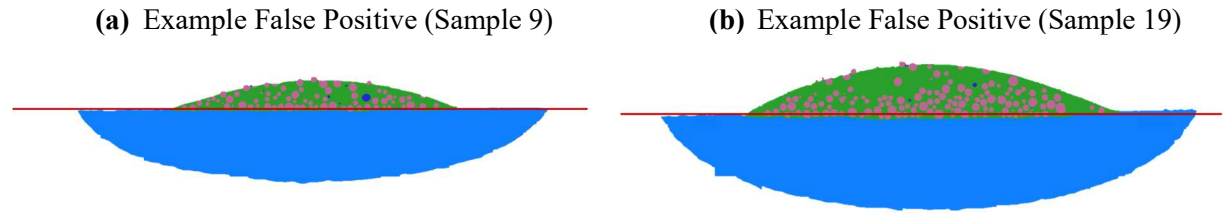

Figure 30: Dilution false positive examples

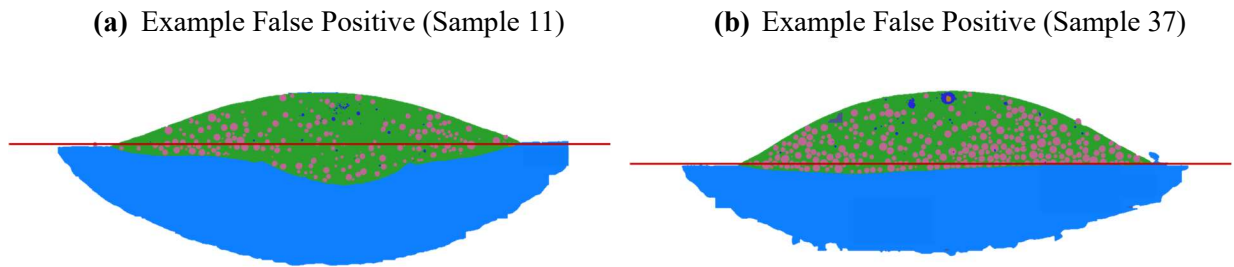

Figure 31: Reinforcement area false positive examples

### 6) Industrial Use Scenario

Figure 32 illustrates a vision-language detection tree that enables fully automated and scalable microstructural classification across diverse manufacturing quality criteria.

Combining VLRs with customized positive/negative prompts, the framework branches through successive assessments such as dilution, HAZ, reinforcement, dissolution, and carbide distribution. At each decision node, cosine similarity scores are normalized (z-scores) and fused across models (e.g., CLIP and FLAVA), enabling robust classification based on multi-modal consensus. This modular architecture allows interactive customization of prompts and aggregation of multiple textual criteria, which can support fine-grained, expert-aligned analysis of segmented microstructures. The final decision tree streamlines high-throughput qualification and reduces manual effort while maintaining expert-level interpretability and control.

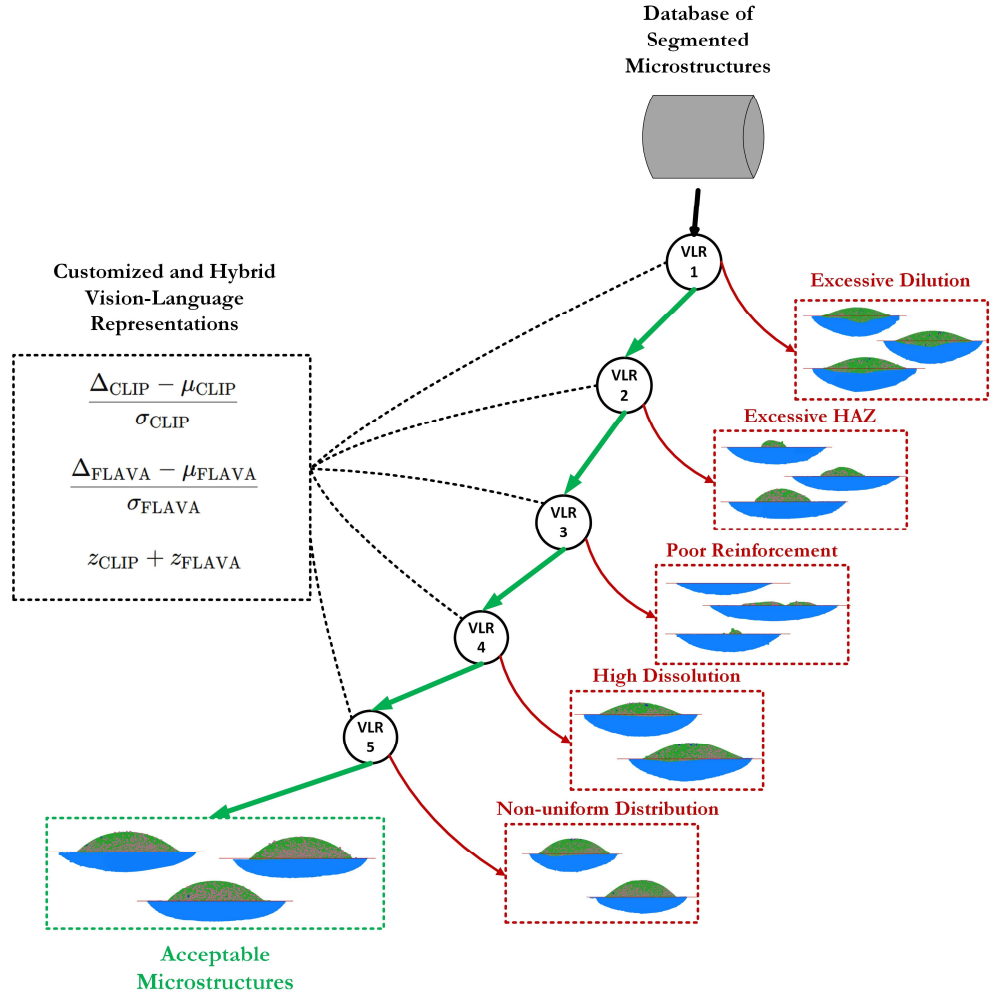

Figure 32: Heterogeneous microstructure detection tree based on customized and hybrid VLRs

A multimodal characterization tool could be realized based on the design presented in Figure 33, with the following functions:

- **Interactive Fusion Strategy Selection:** The interface allows users to choose between vision-language fusion methods such as z-score sum, weighted averaging, or voting. This can enable tailored decision strategies based on application context and confidence calibration needs.
- **Configurable Confidence Thresholding:** Users can adjust the classification threshold in real time using a slider. This can offer transparency and control over sensitivity versus specificity trade-offs during microstructural feature classification.
- **Dual-Model Backend Integration:** By combining embeddings from both FLAVA and CLIP, the tool leverages complementary strengths of image-text alignment. This can improve robustness in classifying complex metallographic phenomena such as reinforcement presence, carbide distribution, and fusion interface.
- **Automated Image Upload and Inference:** The application supports drag-and-drop image input for instant analysis. The backend can automatically compute similarity-based decisions and outputs results alongside confidence scores and textual descriptions.

- **Contextualized Expert Feedback Output:** Besides classification labels, the system retrieves textual descriptions aligned with visual evidence. This behaviour can be accomplished with simpler language models and will simulate expert-style recommendations to support explainable AI in materials qualification workflows.

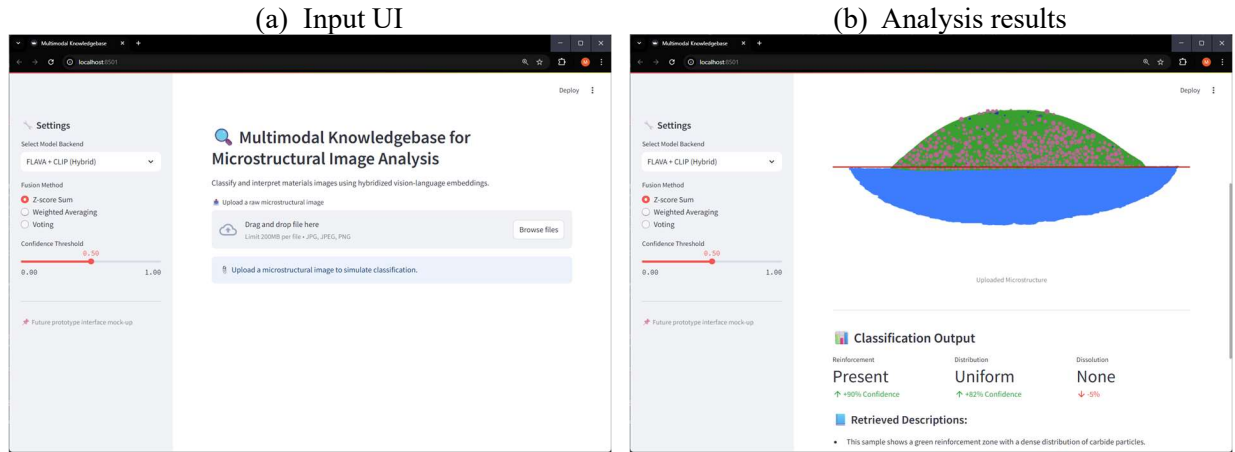

Figure 33: Preliminary design proposal for the development of a multi-modal knowledge base for rapid industrial qualification with customized and hybrid representations as reference. Input interface (a) and output results (b)

## 7) Conclusions and Future Works

This work presents a novel multi-modal framework for linking expert characterization knowledge with heterogeneous microstructural data through customized and hybrid VLRs. Building on the growing potential of foundation models in engineering informatics, we explored and enhanced the capabilities of two state-of-the-art VLMs, CLIP and FLAVA, using a dataset of optical metallographs from additively manufactured MMCs. First, we demonstrated that pre-trained VLMs can serve as viable zero-shot classifiers for complex microstructures by encoding visual and textual representations into a shared embedding space. We then introduced a customized similarity-based scoring approach incorporating multiple positive and negative references to represent expert assessment criteria. Our hybrid scoring method significantly improved classification performance across multiple expert assessments by integrating standardized similarity deltas from CLIP (vision-language) and FLAVA (vision-vision). Second, our analysis revealed the complementary strengths of each model: FLAVA offered sharper visual discrimination, particularly for image-to-image similarity. At the same time, CLIP provided stable text-based alignment even under diverse phrasings and semantic variants. Z-score normalization enhanced decision stability by regularizing the dynamic range of pre-trained modalities within the dataset distribution, enabling accurate classification of samples at the boundary of expert-defined thresholds.

To ensure practical relevance, we also proposed a functional prototype of an industrial characterization knowledge base capable of cross-modal retrieval, semantic reasoning, and expert-guided interpretation. This system is designed with scalability and operator usability and aligns with the DIKW hierarchy. It can be integrated into industrial qualification workflows.

The key findings include:

- Pre-trained models provide sufficient relativity in their latent representations to conduct image-image and image-text similarity analysis

- FLAVA excels in visual similarity alignment, especially when image references are abundant
- CLIP offers stable performance in textual similarity, particularly when phrasing or semantic variation exists
- Hybrid VLRs, enabled by delta scoring and standardization, outperform individual modalities by mitigating overreliance on a single modality representation
- Z-score normalization plays a critical role in shifting borderline decisions and aligning the influence of each modality in the final prediction

There are some limitations of the existing framework and presented experiments. First, the text prompts used in the similarity framework were manually crafted and may not capture the full specificity of expert language. They are also prone to ambiguity when applied to complex or composite features, as highlighted by false positives when classifying bead reinforcement areas. Second, the VLMs used were not yet fine-tuned to exploit the full adaptability to metallographic data. As a result, the models may underperform when faced with out-of-distribution images, inhomogeneous data collection, or anomalous microstructural features. Third, while the hybrid scoring strategy shows promise, its effectiveness depends on careful calibration and may be sensitive to selected references.

Future work will focus on:

- Expanding the framework to incorporate effective expert assessments
- Investigating the generalizability of the hybrid approach to other AM material systems and microscopy modalities (e.g., SEM, XCT)
- Integrating fine-tuning and prompt-engineering capabilities to adapt to evolving characterization criteria

By connecting raw data with the domain expertise using customized VLRs, this study helps establish a more transparent and modular approach to qualifying materials in fast-paced industrial settings. The methodology presented lays the groundwork for building scalable, knowledge-based systems for material qualification. It can be extended to support other heterogeneous materials, imaging modalities, and application domains. Therefore, it lays the groundwork for vision-language-based informatics pipelines that bridge expert intuition and machine reasoning for industrial-scale decision-making.

**Data Availability Statement**

The data that supports the findings of this study will be released upon the publication of this article

**Funding**

This work was supported by the National Research Council Canada (Grant# NRC INT-015-1).

**Disclosure Statement**

The authors declare that they have no known competing financial or non-financial interest that could have appeared to influence the work reported in this paper.

**Author Contribution Statement**

All listed authors meet the criteria for authorship and agree to be accountable for all aspects of this work.

**Mutahar Safdar**: Conceptualization, Methodology, Software, Validation, Formal analysis, Investigation, Writing – original draft. **Gentry Wood**: Resources, Data curation, Writing – original draft. **Max Zimmermann**: Resources, Data curation, Writing – original draft. **Guy Lamouche**: Conceptualization, Methodology, Writing – review & editing, Supervision, Project administration, Funding acquisition. **Priti Wanjara**: Conceptualization, Methodology, Writing – review & editing, Supervision, Project administration, Funding acquisition. **Yaoyao Fiona Zhao**: Conceptualization, Methodology, Writing – review & editing, Supervision, Project administration, Funding acquisition.

Table A1: Expert-labeled metallography dataset

| # | Dilution | Heat Affected Zone | Reinforcement Area | Porosity | Dissolution | Distribution |
|---|---|---|---|---|---|---|
| Sample 1 | reject | reject | accept | reject | reject | reject |
| Sample 2 | accept | accept | accept | reject | accept | reject |
| Sample 3 | reject | accept | accept | reject | accept | reject |
| Sample 4 | reject | accept | accept | reject | reject | reject |
| Sample 5 | accept | accept | accept | accept | accept | accept |
| Sample 6 | reject | accept | accept | reject | accept | reject |
| Sample 7 | accept | accept | accept | accept | accept | accept |
| Sample 8 | accept | reject | accept | accept | reject | reject |
| Sample 9 | accept | reject | accept | reject | accept | reject |
| Sample 10 | accept | accept | accept | reject | reject | accept |
| Sample 11 | reject | reject | accept | reject | reject | reject |
| Sample 12 | reject | reject | accept | reject | reject | reject |
| Sample 13 | accept | accept | accept | reject | accept | reject |
| Sample 14 | reject | accept | accept | reject | reject | reject |
| Sample 15 | accept | reject | reject | accept | accept | accept |
| Sample 16 | accept | reject | accept | accept | accept | reject |
| Sample 17 | accept | reject | accept | accept | accept | accept |
| Sample 18 | accept | reject | accept | accept | accept | reject |
| Sample 19 | accept | accept | accept | accept | reject | reject |
| Sample 20 | accept | reject | accept | reject | accept | reject |
| Sample 21 | reject | accept | accept | reject | accept | reject |
| Sample 22 | reject | accept | accept | reject | reject | reject |
| Sample 23 | accept | accept | accept | reject | reject | accept |
| Sample 24 | accept | accept | accept | reject | accept | reject |
| Sample 25 | accept | reject | reject | accept | accept | accept |
| Sample 26 | accept | accept | accept | reject | reject | accept |
| Sample 27 | accept | accept | accept | accept | accept | accept |
| Sample 28 | accept | accept | accept | accept | accept | reject |
| Sample 29 | accept | reject | accept | accept | accept | reject |
| Sample 30 | accept | reject | accept | accept | accept | reject |
| Sample 31 | accept | accept | accept | reject | accept | reject |
| Sample 32 | NA | NA | reject | NA | NA | NA |
| Sample 33 | reject | accept | accept | reject | reject | reject |
| Sample 34 | reject | accept | accept | reject | reject | reject |
| Sample 35 | accept | accept | accept | accept | reject | accept |
| Sample 36 | reject | reject | accept | accept | accept | reject |
| Sample 37 | reject | accept | accept | reject | accept | reject |
| Sample 38 | reject | reject | accept | reject | reject | reject |
| Sample 39 | accept | accept | accept | accept | accept | accept |
| Sample 40 | accept | accept | accept | accept | accept | reject |